\documentclass[11pt]{article}

\usepackage[margin=1in]{geometry}
\usepackage[table]{xcolor}
\usepackage{algorithm}
\usepackage{amsthm}
\usepackage{hyperref}
\usepackage[normalem]{ulem}
\usepackage{amsmath,amssymb}
\usepackage[mathscr]{euscript} 
\usepackage{tikz} 
\usepackage{graphicx} 
\usepackage{multirow}
\usepackage{subfig}
\usepackage{longtable}
\usepackage{adjustbox}
\usepackage{enumitem}
\usepackage{tabulary}
\usepackage{booktabs}
\newcommand*{\email}[1]{\href{mailto:#1}{\nolinkurl{#1}}} 

\newtheorem{myDefinition}{Definition}
\newtheorem{myTheorem}{Theorem}
\newtheorem{myLemma}{Lemma}
\newtheorem{myCorollary}{Corollary}
\newtheorem{myProposition}{Proposition}
\newtheorem{myExample}{Example}
\newtheorem{myRemark}{Remark}

\newtheorem{myQuestion}{Question}

\newtheorem*{myTheoremNonum}{Theorem}

\newcommand{\bs}[1]{\boldsymbol{#1}}
\newcommand{\bhat}[1]{\boldsymbol{\hat{#1}}}

\def \R{\mathbb{R}}

\def \T{\mathsf{T}}

\def \spn{{\rm span}}

\def \rnk{{\rm rank}}

\def \r{{\rm r}}
\def \Rr{{\rm R}}
\def \d{{\rm d}}
\def \Dd{{\rm D}}
\def \K{{\rm K}}
\def \m{{\rm m}}
\def \M{{\rm M}}
\def \n{{\rm n}}
\def \N{{\rm N}}
\def \q{{\rm q}}
\def \p{{\rm p}}

\def \S{{\mathscr{S}}}
\def \Vstar{{\mathscr{V}}}
\def \sU{\mathscr{U}}
\def \sW{\mathscr{W}}

\def \sV{\mathscr{V}}

\def \I{{\bs{{\rm I}}}}
\def \X{{\bs{{\rm X}}}}

\def \XTp{{\X^{\oell}}}

\def \Y{{\bs{{\rm Y}}}}
\def \A{{\bs{{\rm A}}}}
\def \U{{\bs{{\rm U}}}}

\def \bU{\boldsymbol{{\rm U}}}

\def \bV{\boldsymbol{{\rm V}}}

\def \bB{\boldsymbol{{\rm B}}}

\def \bA{\boldsymbol{{\rm A}}}

\def \O{{\bs{\Omega}}}

\def \Ups{{\bs{\Upsilon}}}

\def \o{{\bs{\omega}}}
\def \ups{{\bs{\upsilon}}}

\def \oell{{\otimes \p}}
\def \otwo{{\otimes 2}}

\def \x{{\bs{\rm x}}}
\def \y{{\bs{\rm y}}}
\def \a{{\bs{{\rm a}}}}
\def \bv{\bs{{\rm v}}}
\def \bw{\bs{{\rm w}}}
\def \bx{\bs{{\rm x}}}
\def \by{\bs{{\rm y}}}
\def \bu{\bs{{\rm u}}}

\def \ba{\bs{{\rm a}}}

\def \i{{\rm i}}
\def \j{{\rm j}}
\def \k{{\rm k}}

\def \Tt{{\rm T}}

\def \ie{\emph{i.e.,~}}
\def \eg{\emph{e.g.,~}}

\newcommand{\rb}{\cellcolor{red}}
\newcommand{\yb}{\cellcolor{yellow}}

\DeclareMathOperator{\supp}{supp}

\newcommand*{\Scale}[2][4]{\scalebox{#1}{\ensuremath{#2}}} 

\usepackage{enumitem}
\setlist[enumerate]{leftmargin=.5in}
\setlist[itemize]{leftmargin=.5in}

\newcommand{\edit}[1]{#1}
\newcommand{\new}[1]{#1}

\usepackage{mathtools}
\DeclarePairedDelimiter{\ceil}{\lceil}{\rceil}

\title{Tensor Methods for Nonlinear Matrix Completion}
\author{
   Greg Ongie\thanks{Mathematical and Statistical Sciences, Marquette University, Milwaukee, WI (\email{gregory.ongie@marquette.edu})}
  \and Daniel Pimentel-Alarc{\'o}n\thanks{Department of Biostatistics, University of Wisconsin, Madison, WI (\email{pimentelalar@wisc.edu})}
  \and Laura Balzano\thanks{Electrical Engineering and Computer Science, University of Michigan, Ann Arbor, MI (\email{girasole@umich.edu})}
  \and Rebecca Willett\thanks{Departments of Statistics and Computer Science, University of Chicago, Chicago, IL (\email{willett@uchicago.edu})}
   \and Robert D.~Nowak\thanks{Department of Electrical and Computer Engineering, University of Wisconsin, Madison, WI (\email{rdnowak@wisc.edu})}
}

\begin{document}

\maketitle

\begin{abstract}
  In the low-rank matrix completion (LRMC) problem, the low-rank
  assumption means that the columns (or rows) of the matrix to be
  completed are points on a low-dimensional linear algebraic variety. This paper extends this thinking to cases where the columns are points on a low-dimensional {\em nonlinear} algebraic variety, a problem we call {\em Low Algebraic Dimension Matrix Completion} (LADMC). Matrices whose columns belong to a union of subspaces are an important special case.
  We propose a LADMC algorithm that leverages existing LRMC methods on a tensorized representation of the data. For example, a second-order tensorized representation is formed by taking the Kronecker product of each column with itself, and we consider higher order tensorizations as well.
  This approach will succeed in many cases where traditional LRMC is guaranteed to fail because the data are low-rank in the tensorized representation but not in the original representation.
  We also provide a formal mathematical justification for the success of our method. In particular, we give bounds of the rank of these data in the tensorized representation, and we prove sampling requirements to guarantee uniqueness of the solution.
  We also provide experimental results showing that the new
  approach outperforms existing state-of-the-art methods for matrix completion under a union of subspaces model.
\end{abstract}

\section{Introduction}
\label{sec:intro}

The past decade of research on matrix completion has shown it is
possible to leverage linear dependencies among columns and/or rows of a
matrix to impute missing values. 
For instance, if each column of a matrix corresponds to a different high-dimensional data point belonging to the same low-dimensional linear
subspace, then the corresponding matrix is low-rank and
missing values can be imputed using low-rank matrix
completion \cite{candes2009exact,candes-tao,recht,recht2010guaranteed,keshavan2010matrix}. 
These ideas continue to impact
diverse applications such as recommender systems \cite{lee2013local}, image inpainting \cite{jin2015annihilating},
computer vision \cite{kennedy2016online}, and array signal processing \cite{sun2015mimo}, among others.

The high-level idea of this body of work is that if the data defining
the matrix has fewer degrees of freedom than
the matrix dimension, that structure provides redundancy that can be
leveraged to impute missing data values. However, the usual low-rank
assumption is not always satisfied in practice. Extending matrix
completion theory and algorithms to exploit low-dimensional nonlinear
structure in data will allow missing data imputation in a far richer
class of problems.

\edit{This paper describes matrix completion in the context of nonlinear algebraic
varieties, a polynomial generalization of linear
subspaces. 
More precisely, let $\X \in \R^{\d \times \N}$ be a matrix whose
columns lie in a low-dimensional algebraic variety $\Vstar \subset \R^\d$.}
Such matrices will be called {\em low
algebraic dimension} (LAD) matrices. 
In the case where $\Vstar$ is a low-dimensional linear variety, \ie a linear subspace, this reduces to low-rank matrix completion (LRMC).
We call the more general
problem of completing LAD matrices {\em low algebraic dimension matrix
completion} (LADMC).

\begin{figure}[htbp]
    \centering
    \includegraphics[width=0.70\textwidth]{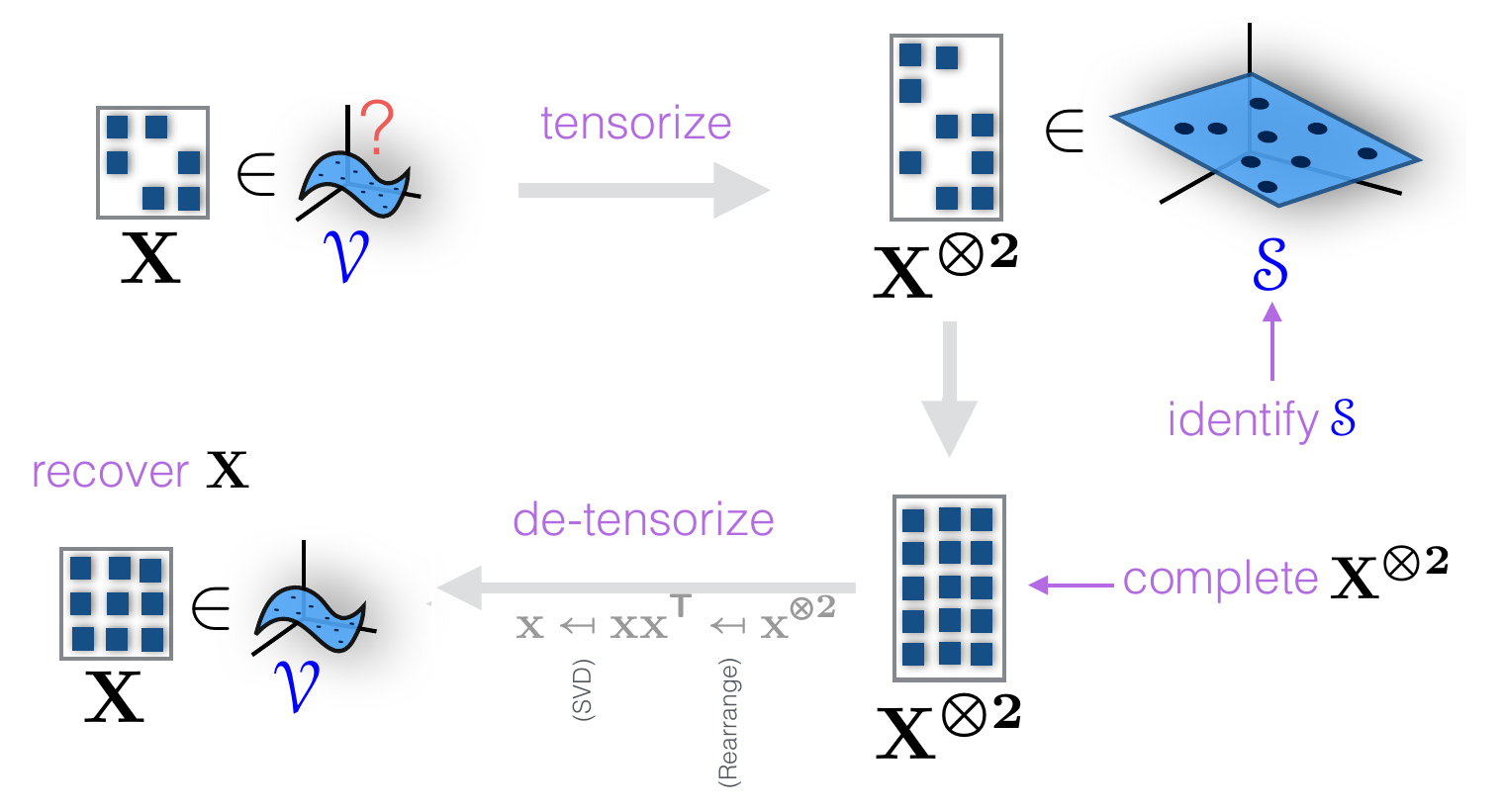}
    \caption{\edit{\textbf{Schematic of the proposed low-algebraic dimension matrix completion (LADMC) approach using a quadratic tensorization.} Given a matrix $\X$ with missing entries we synthesize entries of its tensorized representation $\X^\otwo$, the columnwise Kronecker product of $\X$ with itself. If the columns of $\X$ belong to an unknown variety $\Vstar$ with low algebraic dimension then $\X^\otwo$ will be low-rank, \ie its columns belong to a low-dimensional subspace $\S$. Therefore, assuming the observed entries of $\X^\otwo$ are sufficient to uniquely identify $\S$, it is possible to complete $\X^\otwo$ using standard low-rank matrix completion. Finally, we recover $\X$ by a simple ``de-tensorization'' procedure by applying a rank-one truncated SVD to each reshaped column of $\X^\otwo$.}}
    \label{fig:schematic}
\end{figure}

Recently \cite{ongie} proposed a new LADMC approach based on lifting the
problem to a higher-dimensional representation using
polynomial expansions of the columns of $\X$. The algorithm in
\cite{ongie} can be interpreted as alternating between LRMC in the
lifted representation and {\em unlifting} this low-rank representation
back to the original representation to obtain a completion of the original matrix.  This approach
appears to provide good results in practice, but two problems were
unresolved:
\begin{itemize}
\item
The ``unlifting" step is highly nonlinear and non-convex, and so little can be proved about its accuracy or correctness.
\item
While \cite{ongie} provides an intuitive explanation for the potential of the approach (based on a degrees of freedom argument) and why it may succeed in cases where LRMC fails, a rigorous argument is lacking.
\end{itemize}
\edit{This paper addresses both issues. To address the first point, we propose a new LADMC algorithm based on an explicit tensorization of the data. Specifically, let the \emph{tensorized representation} of a matrix be another matrix where each column is the vectorized $p$-fold Kronecker product of the same column of the original matrix with itself\footnote{\edit{The tensorized representation could also be formally viewed as a $(p+1)$-way tensor, for which low-rank tensor completion techniques could be brought to bear (e.g., \cite{gandy2011tensor}). However, in this work we focus on a matrix flattening of the tensorized representation, since our approach only exploits linear relationships between the columns of the tensorized representation.}}. If the original matrix has missing entries, then we treat an entry in its tensorized representation as missing if it cannot be formed from products of observed entries of $\X$ (\ie if one or more of the factors making up the product is missing in $\X$). Our algorithm consists of three steps: (1) form the tensorized representation of the original matrix with missing entries,
(2) apply standard LRMC to the tensorized representation to recover its missing entries, (3) map each completed column of the tensorized matrix back to the original domain by applying a simple column-wise ``de-tensorization'' step based on a rank-one (truncated) singular value decomposition. Provided the completion in step (2) is correct, then step (3) is guaranteed to recover the correct original column. See Figure \ref{fig:schematic} for a schematic and Algorithm \ref{ladmcAlg} for further details. We present both iterative and non-iterative versions of this algorithm, and experiments showing that its performance is as good or better than state-of-the-art methods in the popular case of the union of subspaces model.}

\edit{To address the second point, we make several contributions towards understanding the information-theoretic requirements of LADMC. We build on recent work that gives deterministic sampling conditions for low-rank matrix completion and subspace clustering with missing entries
\cite{identifiability,LRMCpimentel,infoTheoretic,converse,r2pca}. Specifically, we give conditions guaranteeing the column space of the \emph{tensorized representation} of the data matrix is uniquely identifiable from its \emph{canonical projections}, \ie projections of the subspace onto a collection of canonical basis elements. This sampling model is equivalent to an entry-wise sampling model that assumes sufficiently many data columns are sampled with the same observation pattern. 
The main challenge in our setting is that while data may be missing uniformly at random in the original data matrix $\X$, the pattern of missing data in the tensorized domain is highly structured (not uniformly random) due to the product structure, and in fact only a small fraction of all patterns are possible. Hence, the canonical projections that we may observe in the tensor representation are a restrictive subset of all possible canonical projections. Nonetheless, our main results show that, under mild genericity assumptions on the underlying variety, the admissible canonical projections in the tensor space are indeed sufficient to identify the subspace in the tensorized representation.} 

\edit{To give a flavor of our main results, here we present a simplified version of our necessary and sufficient conditions specialized to the case of a quadratic tensorization and data belonging to a union of subspaces (a special case of an algebraic variety).}

\begin{myTheoremNonum}[Theorem (simplified)] \edit{Assume that the columns of $\X$ belong to a union of $K\geq 4$ subspaces drawn independently and uniformly at random from the Grassmannian of all $r$-dimensional subspaces in $\R^d$. If each column of $\X$ is observed in at least $\ceil*{\sqrt{Kr(r+1)}}+2$ uniformly randomly selected coordinates and the total number of columns is sufficiently large, then unique matrix completion is possible using the proposed LADMC algorithm with a quadratic tensorization. If fewer than $\ceil*{\sqrt{Kr(r+1)}\,}$ coordinates are observed in each column, then unique low rank matrix completion of the quadratic tensorized representation is impossible.}
\end{myTheoremNonum}

\edit{See Corollaries \ref{cor:basicnecc_uos} and \ref{cor:suffuos} for a formal statement of this result and its extension to higher-order tensorizations. These formal statements are in terms of the canonical projections of a subspace, which is equivalent to an entry-wise sampling model that assumes the total number of columns is sufficient to guarantee multiple copies of every observable sampling pattern. Also, we give more general versions of these results in Lemma \ref{lem:basicnecc_gen} and Theorem \ref{thm:suff} that hold for a broader class of algebraic varieties beyond unions of subspaces.}

\edit{Roughly speaking, this result shows that a sampling rate of $m = O(\sqrt{K}r)$ entries per column is necessary and sufficient for unique identification of the column space of the tensorized representation arising from a generic union of $K$ $r$-dimensional subspaces under a quadratic tensorization. In contrast, $m = O(Kr)$ observations per column are necessary and sufficient to uniquely identify the column space of the original matrix, \ie we see an improvement by a factor of $\sqrt{K}$ in the number of observations per column for which LADMC is information-theoretically possible relative to LRMC in the original domain.}

\edit{The rest of the paper lays out our contributions as follows. Section \ref{algorithmSec} presents the algorithm in more detail and Section \ref{sec:algebraicassumptions} describes the algebraic assumptions that are required for the tensorized representation of the data matrix to be low-rank. Then Section \ref{sec:theory} addresses sampling conditions for unique identifiability of the tensorized subspace, giving both necessary (Section \ref{sec:necessary}) and sufficient (Section \ref{sec:sufficient}) conditions, including results that specialize to the union of subspaces model. Section \ref{sec:Exp} presents numerical results.
}

\subsection{Related Work}

Unions of subspaces (UoS) are a special
case of algebraic varieties \cite{ongie,gpca}, and a number of
approaches to matrix completion for a UoS model have been proposed
\cite{mahdithesis,hrmc,infoTheoretic,kGROUSE,ssp14,ewzf,GSSC,waheed,mixture,elhamifar2016high,seversky}; see \cite{manolisreview} for classification and comparison of these approaches for the task of subspace clustering with missing data. Most
these algorithms involve iterating between subspace clustering and completion steps, and relatively little can be guaranteed about their performance. Exceptions include \cite{hrmc} and \cite{seversky}, which give performance guarantees for algorithms based on a non-iterative neighborhood search procedure.  Also, recent work \cite{charles2017subspace,tsakiris18a} gives performance guarantees for a version of the sparse subspace clustering algorithm modified to handle missing data \cite{ewzf}.

Our proposed LADMC approach is closely related to algebraic subspace clustering (ASC), also known as generalized principal component analysis \cite{gpca,yang2005hilbert,ma2008estimation,tsakiris2017algebraic,tsakiris2017filtrated}. Similar to our approach, the ASC framework models unions of subspaces as an algebraic variety, and makes use of tensorizations (\ie Veronese embeddings) of the data to identify the subspaces.
However, the ASC framework has not been extended to the matrix completion setting, which is the main focus of this work. 

\new{A preliminary version of this work was published in a conference proceedings \cite{allerton}. We extend the theory and algorithms in \cite{allerton} to higher order tensorizations of the data matrix; \cite{allerton} only considered quadratic tensorizations. We also correct several issues with the theory in \cite{allerton}. In particular, parts (i) and (ii) of Theorem 2 of \cite{allerton} are incorrect as stated; in this work we correct this result and also extend it to higher order tensorizations (see Corollary \ref{cor:basicnecc_uos}). Additionally, the proof of part (iii) of Theorem 2 in \cite{allerton} is incorrect; here we give a correct proof and likewise extend the result to higher order tensorizations (see Corollary \ref{cor:suffuos}). Finally, we also expand the experiments section by comparing the proposed LADMC algorithm with the previously proposed VMC algorithm \cite{ongie}, and demonstrate the proposed LADMC algorithm for matrix completion of real data.}

\section{Setup and Algorithm}
\label{algorithmSec}

Suppose we observe a subset of the entries of a matrix $$\X = [\x_1,...,\x_\N] \in \R^{\d\times\N}$$ at locations indicated by ones in the binary matrix $\O = [\o_1,...,\o_\N] \in \{0,1\}^{\d\times\N}$. We refer to $\O$ and $\o_i$ as the matrix and vector {\em observation patterns}, respectively.

In traditional low-rank matrix completion one assumes $\X$ is low-rank in order to recover the missing entries. 
We take a different approach in this work. 
Rather than completing $\X$ directly, we consider low-rank completing the \emph{tensorized matrix} 
$$
  \XTp := [\x_1^\oell,...,\x_\N^\oell] \in \R^{\Dd\times\N}.
$$
Here $\x^\oell$ denotes the {\em $\p$-fold tensorization} of a vector $\x$, defined as $\x^\oell := \x \otimes \cdots \otimes \x$ where $\otimes$ is the Kronecker product, and $\x$ appears $\p$ times in the expression. Every tensorized vector $\x^\oell$ can be reordered into a $\p^{\text{th}}$-order symmetric tensor that is uniquely determined by $\Dd := \binom{\d+\p-1}{\p}$ of its entries. For example, the vector $\x^\otwo$ has the same entries as the matrix $\x\x^\T \in \R^{\d\times \d}$, which is uniquely determined by its $\binom{\d+1}{2}$ upper triangular entries. Hence, with slight abuse of notation, we consider tensorized vectors $\x^\oell$ as elements of $\R^\Dd$.

Additionally, given partial observations of matrix $\X$ at locations in $\O$ we can synthesize observations of the tensorized matrix $\XTp$ at all locations indicated by $\O^\oell = [\o_1^\oell,...,\o_\N^\oell]\in \{0,1\}^{\Dd\times\N}$ simply by multiplying the observed entries of $\X$. In particular, if the data column $\x_i$ is observed in $\m$ locations, then the tensorized data column $\x_i^\oell$ can be observed at $\binom{\m+\p-1}{\p}$ locations indicated by ones in the binary vector $\o_i^\oell$. We refer to $\O^\oell$ and $\o_i^\oell$ as the matrix and vector {\em tensorized observation patterns}, respectively.

Remarkably, there are situations where the original data matrix $\X$ is full rank, but the tensorized matrix $\XTp$ is low-rank, owing to (nonlinear) algebraic structure of the data, described in more detail below. 
In these situations, $\XTp$ can potentially be recovered from its entries indicated by $\O^\oell$ using standard low-rank matrix completion algorithms. 

If the LRMC step recovers $\XTp$ correctly, then we can uniquely recover $\X$ from $\X^{\oell}$. To see this, first consider the case of a quadratic tensorization ($\p=2$). Let $\Y = [\y_1,...,\y_\N]$ be the output from LRMC applied to the tensorized matrix. If the completion is correct, \edit{meaning we exactly recover $\y_i = \x_i^{\otwo}$, then we can reshape $\y_i$ into the exactly} rank-1 symmetric $\d\times\d$ matrix $\Y_i = \x_i\x_i^\T$. Hence, we can recover $\x_i$ by the computing the leading eigenvalue-eigenvector pair ($\lambda_i,\bu_i)$ of $\Y_i$ and setting $\x_i = \pm \sqrt{\lambda_i}\bu_i$, where we determine the sign by matching it to the observed entries of $\x_i$.
\edit{If there is noise and the completion is not exact, $\Y_i$ may not be exactly rank-1, but we see empirically that the leading eigenvector provides a denoised estimate.} 
For higher-order tensorizations ($\p \geq 3$), we can recover $\x_i$ from $\y_i$ using a similar procedure: we reshape $\y_i$ into a $\d \times \d^{\p-1}$ and take its rank-one truncated SVD\footnote{\edit{Another approach would be to augment the tensorized representation with the original matrix prior to the LRMC step. Then we could recover $\X$ by restricting the completion to the block corresponding to the original matrix. However, empirically we found this approach to be more sensitive to noise. Also, the proposed de-tensorization step based on a rank-one SVD is crucial to our iterative version of LADMC (see Section \ref{sec:implementation})}.}, \edit{which similarly will only be exact in the noiseless case}.

These observations motivate our proposed algorithm, Low Algebraic Dimension Matrix Completion (LADMC), summarized below in Algorithm \ref{ladmcAlg}; see Figure \ref{fig:schematic} for a schematic.

\begin{algorithm}
\caption{Low Algebraic Dimension Matrix Completion (LADMC).}
\label{ladmcAlg}
\textbf{Input:} Subset of entries of data matrix $\X$.

\textbf{Tensorize:} Form new matrix $\X^{\oell}$ by replacing each column $\x_\i$ with its $\p$-fold tensor product $\x_\i^\oell$ (with missing entries corresponding to any products involving missing entries in $\x_\i$).

\textbf{LRMC:} Let $\Y = [\by_1,...,\by_\N]$ be the low-rank completion of $\XTp$.

\textbf{De-tensorize:} Compute the best rank-one symmetric tensor approximation $\bhat{\x}_i^\oell$ of each column $\y_i$ of $\Y$ such that $\bhat{\x}_i$ matches the observed entries of $\bx_i$.

\textbf{Output:} Completed matrix $\bhat{\X}$ whose $\i^{\rm th}$ column is $\bhat{\x}_\i$.
\end{algorithm}

\subsection{Algebraic variety models and rank of the tensorized matrix}
\label{sec:algebraicassumptions}
Here we describe in more detail the algebraic assumptions that are required for the tensorized data matrix $\XTp$ to be low-rank.

Suppose $\XTp \in \R^{\Dd\times \N}$ is a wide matrix, \ie the number of data columns $\N$ exceeds the tensor space dimension $\Dd$. Then $\XTp$ is rank-deficient if and only if the rows of $\XTp$ are linearly dependent, in which case there exists a vector $\bv \in \R^\Dd$ such that $\bv^\T\x_i^\oell = 0$ for all columns $\x_i$ of $\X$. In other words, the columns $\x_i$ belong to the zero set of the polynomial $q(\x) = \bv^\T\x^{\oell}$. Hence, we have shown the following:

\begin{myProposition}
The tensorized matrix $\XTp \in \R^{\Dd\times \N}$ with $\N\geq\Dd$ is rank deficient if and only if the columns of $\X$ belong to a proper \emph{algebraic variety}, \ie the common zero set of a collection of polynomials that are not all identically zero.
\end{myProposition}

In particular, we focus on the class of varieties defined by \emph{homogeneous polynomials}\footnote{Our approach extends to varieties defined by inhomogenous polynomials if we redefine $\x^\oell$ to be the map $\x \mapsto [\begin{smallmatrix}1\\\x\end{smallmatrix}]^\oell$, \ie augment $\x$ with a $1$ before tensorization.}. A degree-$\p$ homogeneous polynomial is any polynomial of the form $q(\x) = \bv^\T \x^\oell$, for some vector of coefficients $\bv\in \R^\Dd$.
\begin{myDefinition}
A set $\Vstar \subset \R^\d$ is a \emph{(real) projective variety}\footnote{For any homogeneous polynomial $q$ we have $q(\x) = 0$ if and only if $q(\lambda\x) = 0$ for any scalar $\lambda \neq 0$. This means the zero sets of homogeneous polynomials can be considered as subsets of \emph{projective space}, \ie the set of all lines through the origin in $\R^\d$, and this is the source of the term ``projective variety''. For simplicity, we typically consider projective varieties as subsets of Euclidean space $\R^\d$, unless otherwise noted.} if there exist homogeneous polynomials $q_1,...,q_n$ (with possibly different degrees) such that $$\Vstar = \{\x \in \R^\d : q_1(\x) = \cdots = q_n(\x) = 0\}.$$
\end{myDefinition}

An important fact for this work is that a union of subspaces is a projective variety, as shown in the following example.

\begin{myExample}[Unions of subspaces are projective varieties]\label{ex:UoSasvariety}
Suppose $\sU$ and $\sW$ are subspaces of $\R^\d$. Then the union of the two subspaces $\Vstar:= \sU \cup \sW$ is given by the common zero set of the collection of quadratic forms $q_{i,j}(\x) = (\x^\T\bu_i^\perp)(\x^\T\bw_j^\perp)$, where $\{\bu_i^\perp\}$ is a basis of the orthogonal complement of $\sU$ and $\{\bw_j^\perp\}$ is a basis of the orthogonal complement of $\sW$. Hence $\Vstar$ is a projective variety determined by the common zero set of a collection of quadratic forms. More generally, a union of $\K$ distinct subspaces is a projective variety defined by a collection of degree $\K$ polynomials, each of which is a product of $\K$ linear factors; this fact forms the foundation of algebraic subspace clustering methods \cite{ma2008estimation,gpcabook}.
\end{myExample}

Given a matrix whose columns are points belonging to a projective variety, the rank of the associated tensorized matrix is directly related to the dimension of the associated \emph{tensorized subspace}, defined as follows:
\begin{myDefinition}
Let $\Vstar \subset \R^d$ be a projective variety. We define the $\p^{\text{th}}$-order \emph{tensorized subspace} associated with $\Vstar$ by
\begin{equation}
  \S := \spn\{\x^\oell : \x \in \Vstar\} \subset \R^\Dd
\end{equation}
\ie the linear span of all $\p^{\text{th}}$-order tensorized vectors belonging to $\Vstar$.
\end{myDefinition}

If the columns of a matrix $\X$ belong to a projective variety $\Vstar$, then the column space of the tensorized matrix $\XTp$ belongs to the tensorized subspace $\S$, and so the dimension of the tensorized subspace is an upper bound on the rank of the tensorized matrix. In particular, if there are a total of $L$ linearly independent degree $\p$ homogeneous polynomials vanishing on $\Vstar$ then $\S$ is a subspace of $\R^\Dd$ of dimension at most $\Dd-L$. Therefore, if there are sufficiently many such polynomials then $\S$ is a low-dimensional subspace, and hence $\XTp$ is low-rank. \edit{One situation in which this can occur is if any homogeneous polynomial $q(\x)$ of degree $p'$ smaller than the tensor order $p$ vanishes on the variety. In this case, all degree $p$ polynomials given by the product $q(\x)$ by a monomial of degree $p-p'$ also vanish on the variety and are linearly independent, which drives down the dimension $\Rr$ of the tensorized subspace.}

As a more concrete example, consider the special case of a union of subspaces. The linear span of all points belonging to a union of $\K$ $\r$-dimensional subspaces (in general position) defines a subspace of dimension $\min\{\K\r,\d\}$ in $\R^\d$. However, in the tensor space these points lie in a subspace of $\R^\Dd$ whose dimension relative to the tensor space dimension is potentially smaller, as shown in the following lemma.
\begin{myLemma}
\label{basisLem}
Let $\Vstar \subset \R^\d$ be a union of $\K$ $\r$-dimensional subspaces, and let 
$\S \subset \R^\Dd$ be its $\p^{\text{th}}$-order tensorized subspace. Then $\S$ is $\Rr$-dimensional where
\begin{equation}\label{eq:UoSRBnd1}
  \Rr \leq \min\left\{ \K\textstyle\binom{\r+\p-1}{\p},\Dd \right\}.
\end{equation}
\end{myLemma}
The proof of Lemma \ref{basisLem} is elementary: any $\r$-dimensional subspace in $\R^\d$ spans a $\binom{\r+\p-1}{\p}$-dimensional subspace in the tensor space, and so points belonging a union of $\K$, $\r$-dimensional subspaces spans at most a $\K\binom{\r+\p-1}{\p}$-dimensional subspace in the tensor space. 

\edit{Example 1 shows that a union of $K$ subspaces is always defined by a set of polynomials of degree $K$. This suggests that a $K$th-order tensorization is necessary to have small rank in the tensorization domain and hence for successful LRMC. When $K$ is large, this may be computationally prohibitive. However, as we will see, lower tensor orders (\eg $p=2,3$) still allow for completion of data belonging to UoS in scenarios where standard LRMC fails. This is because even though the UoS is always defined by degree $K$ polynomials, several lower degree polynomials also vanish on the UoS, which is sufficient to reduce the rank in the tensorized representation enough to allow for successful LRMC in the tensor domain.} 

More precisely, assuming we need $O(\Rr)$ observations per column to complete a rank-$\Rr$ matrix, completing a matrix whose columns belong to union of $\K$ $\r$-dimensional subspaces in the original space would require $O(\K\r)$ observations per column, but completing the corresponding tensorized matrix would require $O(\K \binom{\r+\p-1}{\p})$ entries (products) per column, which translates to $O({\K}^{1/\p}\r)$ entries per column in the original matrix. This suggests LADMC could succeed with far fewer observations per column than LRMC (\ie $O({\K}^{1/\p}\r)$ versus $O(\K\r)$) in the case of UoS data.

We note that Lemma \ref{basisLem} is a special case of a more general bound due to \cite{chardin1989majoration} that holds for any \emph{equidimensional} projective variety\footnote{The result in \cite{chardin1989majoration} gives an upper bound on the values of the Hilbert function associated with any homogeneous unmixed radical ideal $I \subset k[x_0,...,x_d]$ over a perfect field $k$. We specialize this result to the vanishing ideal of an equidimensional variety in real projective space. In particular, the dimension of the $\p^{\text{th}}$-order tensorized subspace coincides with the Hilbert function of the vanishing ideal evaluated at degree $\p$.}. \edit{Roughly speaking, a projective variety is \emph{equidimensional} if at all points where the variety is smooth the local dimension of the variety (treated as a smooth manifold) is the same}. The bound in \cite{chardin1989majoration} is posed in terms of the \emph{degree} and \emph{dimension} of the variety (see, \eg \cite{cox2007ideals} for definitions of these quantities). Translated to our setting, this result says if $\Vstar$ is a equidimensional projective variety of degree $\K$ and dimension $\r$, then its $\p^{\text{th}}$ order tensorized subspace is $\Rr$-dimensional where $\Rr$ obeys the same bound as in \eqref{eq:UoSRBnd1}. Therefore, given a matrix whose columns belong to an equidimensional projective variety, we should expect that LADMC will succeed with $O({\K}^{1/\p}\r)$ observations per column, where now $\K$ is the degree of the variety and $\r$ is its dimension. In other words, when the data belong to a projective variety with \emph{high degree} and \emph{low dimension} \edit{we expect LADMC may potentially succeed where LRMC may fail}. 


\section{Theory}\label{sec:theory}

\subsection{Limitations of prior theory}
Algorithm \ref{ladmcAlg} is primarily inspired by the ideas in \cite{ongie}. In \cite{ongie}, an informal argument is given for the minimum number of observed entries per data column necessary for successful completion of a tensorized matrix based on the dimension of the corresponding tensorized subspace. Translated to the setting of this paper, the claim made in \cite{ongie} is that in order to successfully complete a matrix $\X$ whose $\p^{\text{th}}$-order tensorized matrix $\XTp$ is rank $\Rr$, we must observe at least $\m_0$ entries per column of $\X$, where $\m_0$ is the smallest integer such that 
\begin{equation}\label{eq:hypnc}
  \textstyle\binom{\m_0+\p-1}{\p} > \Rr,
\end{equation}
\ie unique low-rank completion ought to be possible when the number of observations per column of the tensorized matrix exceeds its rank. This conjecture was based on the fact that $\Rr+1$ is the necessary minimum number of observations per column to uniquely complete a matrix whose columns belong to a $\Rr$-dimensional subspace in general position \cite{converse}. Additionally, $\Rr+1$ observations per column is sufficient for unique completion assuming there are sufficiently many data columns and the observation patterns are sufficiently diverse \cite{LRMCpimentel}. \edit{A similar argument is given in \cite{fan2020polynomial}, where the authors identify the minimum number of parameters needed to uniquely determine a matrix $\X$ among the set of all matrices whose tensorization has the same rank, \ie the set $\{\X : \R^{\d\times \N} : \rnk(\X^\oell) = \Rr\}$. This number of parameters is conjectured to be the necessary minimum number of observed entries necessary to recover $\X$, which matches the condition in \eqref{eq:hypnc} when considering the sampling model in this work.}

However, there are two key technical issues not considered in \cite{ongie} \edit{and \cite{fan2020polynomial}} that prevent their arguments from being rigorous. One is related to the fact that the patterns of missing entries in the tensorized matrix are highly structured due to the tensor product. Consequently, the set of realizable observation patterns in the tensorized matrix is severely limited. These constraints on the observation patterns imply that existing LRMC theory (which typically requires uniform random observations) does not apply directly to tensorized representations. 

The other technical issue not considered by \cite{ongie} \edit{and \cite{fan2020polynomial}} is that the tensorized subspace (\ie the column space of the tensorized matrix) is \emph{not} always in general position as a subspace of $\R^\Dd$. For example, if an $\Rr$-dimensional subspace is in general position then the restriction of the subspace to any subset of $\Rr$ canonical coordinates is $\Rr$-dimensional (\ie if $\bB \in \R^{\Dd\times \Rr}$ is any basis matrix for the subspace, then all $\Rr\times \Rr$ minors of $\bB$ are non-vanishing). However, generally this property does not hold for tensorized subspaces arising from union of subspaces, even if the subspaces in the union are in general position (see Example \ref{ex:counter2} below). General position assumptions are essential to results that describe deterministic conditions on the observation patterns allowing for LRMC \cite{identifiability,infoTheoretic}. Hence, the direct application of these results to the LADMC setting is not possible.

For these reasons it was unclear whether unique completion via LADMC was information-theoretically possible. In fact, we prove there are cases where condition \eqref{eq:hypnc} is satisfied, but where $\XTp$ cannot be completed uniquely using LRMC, even with an unlimited amount of data (see Example \ref{ex:counter1} below). In the remainder of this section we derive necessary and sufficient conditions under which unique completion via LADMC is possible, and compare these with condition \eqref{eq:hypnc}.

\subsection{Unique identifiability of the tensorized subspace}
To simplify our results, we consider a sampling model in which we observe exactly $\m$ entries per column of the original matrix.
The main theoretical question we are interested in is the following:

\begin{myQuestion}\label{Q1}
What is the minimum 
number of observations per column, $\m$, of the original matrix for which unique completion is \emph{information-theoretically possible} with Algorithm \ref{ladmcAlg}?
\end{myQuestion}

Rather than study Question~\ref{Q1} directly, we will study the more basic problem of the unique identifiability of the tensorized subspace (\ie the column space of the tensorized matrix) from its projections onto subsets of canonical coordinates.
This is related to Question 1 as follows: Suppose that we observe multiple columns of the original matrix $\X$ with the same observation pattern. Then we will observe the corresponding columns of the tensorized matrix $\XTp$ with the same tensorized observation pattern. Hence, given sufficiently many columns that are in general position, we can compute a basis of the projection of the tensorized subspace onto coordinates specified by the tensorized observation pattern. This means that given sufficiently many data columns observed with observation patterns of our choosing, we could in principle compute any projection of the tensorized subspace onto coordinates specified by any tensorized observation pattern. Hence, we consider instead the following closely related question:

\begin{myQuestion}\label{Q2}
What is the minimum value of $\m$ for which the tensorized subspace is \emph{uniquely identifiable} from its projections onto all possible tensorized observation patterns arising from a sampling of $\m$ entries per column in the original domain?
\end{myQuestion}

To more precisely describe what we mean by \emph{unique identifiability} of the tensorized subspace in Question 2, we introduce the following notation and definitions. 

For any observation pattern $\o \in \{0,1\}^\d$ we let $|\o|$ denote the total number of ones in $\o$. We say the tensorized observation pattern $\ups = \o^\oell$ is of size $\m$ if $|\o| = \m$. Note that if $\ups$ is a tensorized observation pattern of size $\m$, then $\ups$ has $\binom{\m+\p-1}{\p}$ ones, \ie $|\ups| = \binom{\m+\p-1}{\p}$. 
For any observation pattern $\ups \in \{0,1\}^\Dd$ and any vector $\y \in \R^\Dd$ let $\y_\ups \in \R^{|\ups|}$ denote the restriction of $\by$ to coordinates indicated by ones in $\ups$. Likewise, for any subspace $\S\subset\R^{\Dd}$ we let $\S_\ups \subset \R^{|\ups|}$ denote the subspace obtained by restricting all vectors in $\S$ to coordinates indicated by ones in $\ups$, and call $\S_\ups$ the \emph{canonical projection} of $\S$ onto $\ups$.
For any subspace $\S \subset \R^\Dd$ and any observation pattern matrix $\Ups = [\ups_1 \ \ldots \ \ups_\n] \subset \{0,1\}^{\Dd\times \n}$ we define $\mathcal{S}(\S,\Ups)$ to be the set of all subspaces $\S'$ whose canonical projections onto observation patterns in $\Ups$ agree with those of $\S$, \ie all $\S'$ such that $\S'_{\ups_\i} = \S_{\ups_\i}$ for all $\i=1, \ldots, \n$. We say a subspace $\S$ is \emph{uniquely identifiable} from its canonical projections in $\Ups$ if $\mathcal{S}(\S,\Ups) = \{\S\}$.

To aid in determining whether a subspace is uniquely identifiable from a collection of canonical projections, we introduce the \emph{constraint matrix} $\bA = \bA(\S,\Ups)$, defined below. 
\begin{myDefinition}\label{def:cmatrix} Given a subspace $\S \subset\R^\Dd$ and observation pattern matrix $\Ups = [\ups_1,...,\ups_n] \in \{0,1\}^{\Dd\times \n}$, define the \emph{constraint matrix} $\bA \in \R^{\Dd\times \Tt}$ as follows: for all $i=1,...,\n$ suppose $\M_i : = |\ups_i|$ is strictly greater than $\Rr'_i := \dim \S_{\ups_i}$, and let $\bA_{\ups_i} \in \R^{\M_i \times (\M_i-\Rr_i')}$ denote a basis matrix for $(\S_{\ups_i})^\perp \subset \R^{\M_i}$, the orthogonal complement of the canonical projection of $\S$ onto $\ups_i$, so that $\ker \bA_{\ups_i}^\T = \S_{\ups_i}$. Define $\bA_i \in \R^{\Dd \times \edit{(\M_i-\Rr'_i)}}$ to be the matrix whose restriction to the rows indexed by $\ups_i$ is equal to $\bA_{\ups_i}$ and whose restriction to rows not in $\ups_i$ is all zeros. Finally, set $\bA = [\bA_1~\ldots~\bA_n]$, which has a total of $\Tt = \sum_{i=1}^n (\M_i-\Rr'_i)$ columns.
\end{myDefinition}

The intuition here is that the orthogonal complement of each $\S_{\ups_i}$ constrains the set of subspaces consistent with the observed projections, and $\bA$ reflects the collection of these constraints across all $\n$ observation patterns. The following result shows that unique identifiability of a subspace from its canonical projections is equivalent to a rank condition on the corresponding constraint matrix:

\begin{myLemma}\label{lem1}
An $\Rr$-dimensional subspace $\S$ is uniquely identifiable from canonical projections in $\Ups$ if and only if $\dim \ker \bA^\T = \Rr$, in which case $\S = \ker \bA^\T$.
\end{myLemma}
\begin{proof}
By construction, $\S' \in \mathcal{S}(\S,\Ups)$ if and only if $\S'_{\ups_i} = \S_{\ups_i} = \ker \A_{\ups_i}^\T$ for all $i = 1,...,\n$ if and only if $\S' \in \ker \bA^\T$. Hence, the set $\mathcal{S}(\S,\Ups)$ coincides with all $\Rr$-dimensional subspaces contained in $\ker \A^\T$. In particular, we always have $\S \subset \ker \bA^\T$ and by linearity, $\spn\{ \x \in \S : \S\in\mathcal{S}(\S,\Ups)\} = \ker\A^\T$. Hence, if ${\dim \ker\bA^\T = \Rr}$ it must be the case that $\ker\bA^\T = \S$.
\end{proof}

\begin{myRemark}
Lemma \ref{lem1} gives an empirical criterion for determining whether a subspace is uniquely identifiable: given canonical projections of a subspace $\S$, one can construct the constraint matrix $\bA$ above and numerically check if the dimension of the null space of $\bA^\T$ agrees with the subspace dimension $\Rr$, if it is known. We will use this fact to explore the possibility of unique identifiability of tensorized subspaces arising from unions of subspaces of small dimensions (see Table \ref{fig:tables}).
\end{myRemark}

\subsection{Generic unions of subspaces}
We are particularly interested in understanding Question 2 in the context of tensorized subspaces arising from a \emph{union of subspaces} (UoS), \ie varieties of the form $\Vstar = \cup_{k=1}^\K \sU_k$, where each $\sU_k \subset \R^\d$ is a linear subspace. To simplify our results, we will focus on UoS where each subspace in the union has the same dimension $\r$. We will also often make the assumption that the UoS is \emph{generic} \edit{in an algebro-geometric sense (see, \eg \cite{carlini2012subspace})}. \edit{More precisely, we say a result holds for a \emph{generic union of $\K$ $\r$-dimensional subspaces in $\R^\d$} if it holds for all  collections of subspaces $(\sU_1,...,\sU_\K)$ belonging to an (often unspecified) open dense subset of the product space $\mathbb{G}(\r,\R^\d) \times \cdots \times \mathbb{G}(\r,\R^\d)$, where $\mathbb{G}(\r,\R^\d)$ denotes the Grassmannian of all $\r$-dimensional subspaces in $\R^\d$.} \edit{ Put in probabilistic terms, a property holds for a generic UoS if and only if it holds with probability one for any union of random subspaces whose joint distribution is absolutely continuous with respect to the $K$-fold product of the uniform measure on $\mathbb{G}(\r,\R^\d)$. In particular, if a property holds for a generic union of $\K$ $r$-dimensional subspaces then it holds with probability one for a union of $K$ random subspaces drawn independently from the uniform measure on $\mathbb{G}(\r,\R^\d)$.} 

We will repeatedly make use of the following facts regarding generic UoS (see, \eg \cite{yang2005hilbert}):

\begin{myProposition}\label{prop:genericuos}
Let $\S$ be the $p$th order tensorized subspace arising from a generic union of $\K$ $\r$-dimensional subspaces in $\R^\d$. Then $\Rr(\d,\K,\r,\p) := \dim \S$ is a constant that depends only on $\d,\K,\r,\p$.
\end{myProposition}
Treated as a function of $\p$, the quantity $\Rr(\d,\K,\r,\p)$ is called the \emph{Hilbert function} of a generic UoS (also called a ``generic subspace arrangement''), and is studied in \cite{yang2005hilbert,carlini2012subspace,derksen}.

\subsection{Necessary conditions for unique identifiability of tensorized subspaces}
\label{sec:necessary}

\edit{Here we derive necessary conditions for unique identifiability of tensorized subspaces from their canonical projections based on properties of the corresponding constraint matrix (see Definition \ref{def:cmatrix}). In Lemma \ref{lem:basicnecc_gen} we give a general result that holds for tensorized subspaces arising from any projective variety then specialize to the case of those arising from generic UoS in Corollary \ref{cor:basicnecc_uos}. In particular, we show that there are conditions under which the sampling rate \eqref{eq:hypnc} conjectured to be necessary and sufficient in \cite{ongie} is indeed necessary, but also demonstrate that it fails to be necessary and sufficient in general (see Examples \ref{ex:counter1} and \ref{ex:counter2}).}

First, observe that Lemma \ref{lem1} implies a general necessary condition for unique identifiability of an $\Rr$-dimensional tensorized subspace $\S \subset \R^\Dd$: in order for $\dim \ker \bA^\T = \Rr$ the number of columns $\bA$ needs to be at least $\Dd-\Rr$, simply by considering matrix dimensions. This immediately gives the following result.

\begin{myLemma}\label{lem:basicnecc_gen}
Let $\sV \subset \R^\d$ be a projective variety whose $\p^{\text{th}}$-order tensorized subspace $\S \subset \R^\Dd$ is $\Rr$-dimensional. Suppose we observe canonical projections of $\S$ onto $\n$ distinct tensorized observation patterns $\Ups = [\ups_i,...,\ups_n]\subset \{0,1\}^{\Dd\times n}$. For all $i=1,...,n$ define $\M_i := |\ups_i|$ and $\Rr_i' := \dim \S_{\ups_i}$. Then a necessary condition for $\S$ to be uniquely identifiable is
\begin{equation}\label{eq:basicnecc_gen}
  \sum_{i=1}^n(\M_i-\Rr'_i) \geq \Dd - \Rr.
\end{equation}
\end{myLemma}

Lemma \ref{lem:basicnecc_gen} has several implications regarding the necessary sample complexity for tensorized subspaces arising from a union of subspaces. Consider the case where $\S$ is the $\p^{\text{th}}$-order tensorized subspace corresponding to a \emph{generic} union of $\K$ subspaces of dimension $\r$. Suppose $\Ups$ consists of all $\binom{\d}{\m}$ tensorized observation patterns of size $\m$, \ie each column $\ups_i$ of $\Ups$ has $\M = \binom{\m+\p-1}{\p}$ ones. From Lemma \ref{lem1} we know that $\dim \S_{\ups_i} = \Rr'$ where $\Rr' \leq \Rr \leq \K\binom{\r + \p-1}{\p}$ and where the value of $\Rr'$ is the same for all tensorized observation patterns $\ups_i$ by genericity \edit{(for more on this point see the discussion following Theorem \ref{thm:suff} below)}. This means the constraint matrix has a total of $\binom{\d}{\m}(\M-\Rr')$ columns, which gives the following necessary condition for unique identifiability of tensorized subspaces arising from generic UoS:

\begin{myCorollary}\label{cor:basicnecc_uos}
Let $\sV \subset \R^\d$ be a generic union of $\K$ $\r$-dimensional subspaces. Suppose its $\p^{\text{th}}$-order tensorized subspace $\S \subset \R^\Dd$ is $\Rr$-dimensional. Let $\Rr' \leq \Rr$ be the dimension of $\S$ projected onto any tensorized observation pattern of size $\m$. Then a necessary condition for $\S$ to be uniquely identifiable from its canonical projections onto all possible tensorized observation patterns of size $\m$ is
\begin{equation}\label{eq:basicnecc}
  \textstyle\binom{\d}{\m}\left(\M-\Rr'\right) \geq \Dd - \Rr.
\end{equation}
where $\M = \binom{\m+\p-1}{\p}$ and $\Dd = \binom{\d+\p-1}{\p}$.
\end{myCorollary}

Immediately from \eqref{eq:basicnecc}, we see that a simpler, but weaker, necessary condition for unique identifiability is $\M > \Rr'$, which is independent of the ambient dimension $\d$. In fact, assuming $\m > \p$ and the ambient dimension $\d$ is sufficiently large, then the condition in \eqref{eq:basicnecc} reduces to $\M > \Rr'$. To see this, observe that $\binom{\d}{\m} = O(\d^\m)$ and $\Dd = \binom{\d+\p-1}{\p} = O(\d^\p)$ and so $(\Dd-\Rr)/\binom{\d}{\m} < 1$ for large enough $\d$. Hence, in this case \eqref{eq:basicnecc} reduces to $\M > \Rr'$. In the event that $\Rr' = \Rr$, this further reduces to the condition $\m \geq \m_0$ given in \eqref{eq:hypnc}, the rate conjectured to be necessary and sufficient in \cite{ongie}. 

However, the following two examples show that when some of the above assumptions are violated (\eg when $\m\leq \p$ or $\Rr'< \Rr$) the condition given in \eqref{eq:hypnc} is neither necessary nor sufficient for unique recovery of the tensorized subspace.

\begin{myExample}\label{ex:counter1}
Suppose $\Vstar$ is a generic union of two 1-D subspaces under a quadratic tensorization ($\K =2, \r = 1, \p = 2$). 
Suppose we consider all tensorized observation patterns of size $\m = 2$. In this case we have $\M = 3 > 2 = \Rr' = \Rr$, which satisfies the condition \eqref{eq:hypnc}. Yet, the necessary condition \eqref{eq:basicnecc} is violated in all ambient dimensions $\d \geq 3$ since
\begin{equation}
{\textstyle \binom{\d}{2}}\cdot \underbrace{1}_{\M - \Rr'} < \underbrace{\textstyle\binom{\d+1}{2}}_{\Dd}\ \ - \underbrace{2}_{\Rr}.
\end{equation}
Hence, unique identifiability of the tensorized subspace is impossible in this case, which shows condition \eqref{eq:hypnc} is not sufficient. However, if we increase the number of observations per column to $\m = 3$, it is easy to show the necessary condition \eqref{eq:basicnecc} is always met in dimensions $\d \geq 4$, and experimentally we find that the sufficient condition $\dim \ker \A^\T = \Rr$ of Lemma \ref{lem1} is also met (see Table \ref{fig:tables}).
\end{myExample}

\begin{myExample}\label{ex:counter2}
\new{
Suppose $\Vstar$ is a generic union of two 2-D subspaces under a quadratic tensorization in 4-dimensional ambient space ($\K =2, \r = 2, \p =2, \d=4$). Suppose we consider all observation patterns of size $\m = 3$. In this case $\M = 6 = \Rr$, which violates condition \eqref{eq:hypnc}. However, we have $\Rr' = 5$ since the canonical projection of the tensorized subspace onto a tensorized observation pattern of size $\m = 3$ has the same dimension as a tensorized subspace arising from a generic union of two 2-D subspaces in $\R^3$, which has dimension $5$. Hence, the necessary condition \eqref{eq:basicnecc} is satisfied:
\begin{equation} 4= 
\textstyle\binom{4}{3}\cdot \underbrace{1}_{\M - \Rr'} = \underbrace{\textstyle\binom{5}{2}}_{\Dd}\ \ - \underbrace{6}_{\Rr} = 4.
\end{equation}
 This shows that unique identificaiton of the tensorized subspace may still be possible in this case. In the supplementary materials we prove that the sufficient condition $\dim \ker \A^\T = \Rr$ of Lemma \ref{lem1} holds in this case, which shows the tensorized subspace is uniquely identifiable. Therefore, condition \eqref{eq:hypnc} is not always necessary.
}
\end{myExample}

\begin{table*}[ht!]
\centering
{\footnotesize \textbf{$2$nd order tensorization ($\p=2$)}\\[1em]
\begin{minipage}{0.32\textwidth}
\centering
\textbf{smallest $\m$ s.t.~$\binom{\m+1}{2} > \Rr$}\\[0.5em]
\begin{tabular}{r|llllllll}
K\textbackslash{}r & 1 & 2 & 3 & 4 & 5\\
\hline
1 & 2 & 3 & 4 &  5 &  6\\
2 & \rb 2 & \yb 4 & \yb 5 &  \yb 6 & \yb 8\\
3 & 3 & 4 & 6 &  \yb 8 & \yb 10\\
4 & 3 & 5 & 7 &  9 & 11\\
5 & 3 & 6 & 8 & 10 & 12
\end{tabular}\\[1em]
\end{minipage}
\centering
\begin{minipage}{0.32\textwidth}
\centering
\textbf{smallest $\m$ s.t.~\eqref{eq:basicnecc} holds}\\[0.5em]
\begin{tabular}{r|llllllll}
K\textbackslash{}r & 1 & 2 & 3 & 4 & 5\\
\hline
1 & 2 & 3 & 4 &  5 &  6\\
2 &  3  & 3  &  4  &   5  &  6 \\
3 & 3 & 4 & 6 &   7 &  9 \\
4 & 3 & 5 & 7 &  9 &  11\\
5 & 3 & 6 & 8 & 10 &  12
\end{tabular}
\end{minipage}
\begin{minipage}{0.32\textwidth}
\textbf{smallest $\m$ s.t.~$\rm{dim} \ker \bA^\T = \Rr$}\\[0.5em]
\begin{tabular}{r|llllllll}
K\textbackslash{}r & 1 & 2 & 3 & 4 & 5\\
\hline
1 & 2 & 3 & 4 &  5 &  6\\
2 &  3  &  3  &  4  &   5  &   6 \\
3 & 3 & 4 & 6 &   7 &   9 \\
4 & 3 & 5 & 7 &  9 &  11\\
5 & 3 & 6 & 8 & 10 &  12
\end{tabular}
\end{minipage}\\[1em]

\textbf{$3$rd order tensorization ($\p=3$)}\\[0.5em]
\begin{minipage}{0.32\textwidth}
\centering
\textbf{smallest $\m$ s.t. $\binom{\m+2}{3}>\Rr$}\\[0.5em]
\begin{tabular}{r|llllllll}
K\textbackslash{}r & 1 & 2 & 3 & 4 & 5 & 6 \\
\hline
1 & \rb 2 & 3 & 4 & 5 & 6  & 7 \\
2 & \rb 2 & 3 & \yb 5 & \yb 6 & \yb 7 & \yb 8 \\
3 & \rb 2 & 4 & \yb 5 & \yb 7 & \yb 8  & \yb 10\\
4 & 3 & 4 & \yb 6 & 7 & \yb 9  & \yb 11\\
5 & 3 & 5 & 6 & 8 & 10 & 11\\
6 & 3 & 5 & 7 & 9 & 10 & 12\\
7 & 3 & 5 & 7 & 9 & 11 & 12
\end{tabular}\\[0.5em]
\end{minipage}
\begin{minipage}{0.32\textwidth}
\centering
\textbf{smallest $\m$ s.t. \eqref{eq:basicnecc} holds}\\[0.5em]
\begin{tabular}{r|llllllll}
K\textbackslash{}r & 1 & 2 & 3 & 4 & 5 & 6 \\
\hline
1 &  3 & 3 & 4 & 5 & 6  & 7  \\
2 &  3 & 3 &  4 &  5 &  6  &  7  \\
3 &  3 & 4 &  4 &  5 &  6  &  7  \\
4 & 3 & 4 &  5 & 7 &  8 &  9 \\
5 & 3 & 5 & 6 & 8 & 10 &  11\\
6 & 3 & 5 & 7 & 9 & 10 & 12 \\
7 & 3 & 5 & 7 & 9 & 11 & 12 
\end{tabular}
\end{minipage}
\begin{minipage}{0.32\textwidth}
\centering
\textbf{smallest $\m$ s.t. $\dim \ker \A^\T = \Rr$}\\[0.5em]
\begin{tabular}{r|llllllll}
K\textbackslash{}r & 1 & 2 & 3 & 4 & 5 & 6 \\
\hline
1 &  3 & 3 & 4 & 5 & 6  & 7  \\
2 &  3 & 3 &  4 &  5 &  6  &  7  \\
3 &  3 & 4 &  4 &  5 &  6  &  7  \\
4 & 3 & 4 &  5 & 7 &  8 &  9 \\
5 & 3 & 5 & 6 & 8 & 10 &  11\\
6 & 3 & 5 & 7 & 9 & 10 & 12 \\
7 & 3 & 5 & 7 & 9 & 11 & 12 
\end{tabular}
\end{minipage} }
\caption{\small \textbf{Evidence that necessary condition \eqref{eq:basicnecc} is also sufficient for unique identification of tensorized subspaces.} Here we identify the minimal value of $\m$ for which the tensorized subspace arising from a union of $\K$, $\r$-dimensional generic subspaces is uniquely identifiable from its canonical projections onto all possible tensorized observations patterns of size $\m$. The left-most table gives the smallest value of $\m$ satisfying condition \eqref{eq:hypnc} that was conjectured to be necessary and sufficient in \cite{ongie}. The middle table reports the smallest value of $\m$ satisfying the necessary condition \eqref{eq:basicnecc}. The right-most table reports the smallest value of $\m$ satisfying the sufficient condition $\ker \bA^\T = \Rr$ given in Lemma \ref{lem1}, which is verified numerically by constructing the constraint matrix $\bA$ from a randomly drawn UoS. The middle and right-most tables are the same, showing the necessary condition \eqref{eq:basicnecc} is also sufficient in these cases. In the left-most tables, red boxes indicate values less than the true necessary and sufficient $\m$, and yellow indicates values more than the true necessary and sufficient $\m$, illustrating the shortcomings of previous theory that have been addressed in this paper.}
\label{fig:tables}
\end{table*}

A natural question is whether the necessary condition in Corollary \ref{cor:basicnecc_uos} is also sufficient, \ie if \eqref{eq:basicnecc} holds do we have unique identifiability of the tensorized subspace? Table \ref{fig:tables} shows the results of numerical experiments that suggest this is indeed the case. In particular, we generated a generic UoS in ambient dimension $\d=12$ for a varying number of subspaces and their dimension, computed their tensorized subspace, and constructed the constraint matrix $\bA$ from all possible canonical projections of the tensorized subspace onto tensorized observation patterns of size $\m$. Then we searched for the minimal value of $\m$ for which the necessary and sufficient condition $\dim \ker\bA^\T = \Rr$ given in Lemma \ref{lem1} holds\footnote{If the condition $\ker\bA^\T = \Rr$ holds for one random realization of a union of $\K$ $\r$-dimensional subspaces, then it holds generically since the condition $\ker\bA^\T = \Rr$ can be recast as a polynomial system of equations in terms of the entries of a collection of basis matrices for each subspace in the UoS.}. We compare this with the minimum value of $\m$ for which the necessary condition \eqref{eq:basicnecc} holds, and we found they agree in all cases considered.

Given the strong numerical evidence, we conjecture that the necessary condition \eqref{eq:basicnecc} is also sufficient. While we do not prove this conjecture in this work, in the next section we give a sufficient condition that is only slightly stronger than the necessary condition \eqref{eq:basicnecc} and orderwise optimal in terms the number of subspaces and their dimension in many cases.

\subsection{Sufficient conditions for unique identifiability of tensorized subspaces}
\label{sec:sufficient}

\edit{In this section we identify conditions under which canonical projections onto \emph{all possible} tensorized observation patterns of arising from a sampling of $\m$ coordinates in the original domain are sufficient for unique identification of the tensorized subspace. In Theorem \ref{thm:suff} we state a general result that holds for tensorized subspaces arising from any projective variety, which we later specialize to generic UoS in Collorary \ref{cor:suffuos}. In particular, Collorary \ref{cor:suffuos} relates the sufficient sampling rate in Theorem \ref{thm:suff} to properties of the \emph{Hilbert function} of a generic UoS, a tool from algebraic geometry. Proposition \ref{prop:hilbert} summarizes known results regarding the Hilbert function of generic UoS, which allows us to establish explicit sufficient sampling rates under certain restrictions on the number of subspaces, their dimension, and the tensorization order. }

First, we state our main result which holds for any projective variety:
\begin{myTheorem}\label{thm:suff}
Let $\sV \subset \R^\d$ be a projective variety whose $\p^{\text{th}}$-order tensorized subspace $\S$ is $\Rr$-dimensional. Suppose there exists a tensorized observation pattern $\ups = \o^\oell$ such that $|\ups| > \Rr$ and $\dim \S_\ups = \Rr$. Then $\S$ is uniquely identifiable from its canonical projections onto all possible tensorized observation patterns of size $\m \geq |\o| + \p$.
\end{myTheorem}
We give the proof of Theorem \ref{thm:suff} in Appendix \ref{sec:app:thm1}.
Roughly speaking, Theorem \ref{thm:suff} says that a sampling rate of $\m \geq |\o| + \p$ (\ie $\m$ observed entries per data column of the original matrix) is sufficient to ensure unique LADMC is information-theoretically \emph{possible} (given sufficiently many columns and sufficiently diverse observation patterns). Note that Theorem \ref{thm:suff} does not make any general position assumptions about the tensorized subspace.

\new{By specializing to the case of tensorized subspaces generated by generic UoS, we are able to more explicitly characterize the sampling rate appearing in Theorem \ref{thm:suff}.}

\new{Consider the tensorized subspace $\S$ of a generic union of $\K$ $\r$-dimensional subspaces $\Vstar \subset \R^\d$. Recall that we define $\Rr(\d,\K,\r,\p) = \dim \S$ , \ie the dimension of the tensorized subspace depends only on $\d,\K,\r,\p$ (see Proposition \ref{prop:genericuos}). Now, given any tensorized observation pattern $\ups = \o^\oell$ of size $\m^*$, observe that $\S_\ups$ is equal to the tensorized subspace arising from $\Vstar_\o \subset \R^{\m^*}$, the UoS restricted to the $\m^*$ coordinates specified by $\o$. Provided $\m^* > \r$, $\Vstar_\o$ is again a generic union of $\K$ $\r$-dimensional subspaces except now in $\m^*$-dimensional ambient space. Hence, $\S_\ups$ has the same dimension as the tensorized subspace arising from a generic UoS in $\m^*$-dimensional ambient space, and so we have $\dim \S_\ups = \Rr(\m^*,\K,\r,\p)$ \emph{for any} tensorized observation pattern $\ups$ of size $\m^* >\r$. This fact combined with Theorem \ref{thm:suff} gives the following immediate corollary.}
\begin{myCorollary}\label{cor:suffuos}
\new{
Let $\sV \subset \R^\d$ be a generic union of $\K$ $\r$-dimensional subspaces and let $\S \subset \R^\Dd$ be its $\p^{\text{th}}$-order tensorized subspace.
Assume $\m^*$ is such that $r < \m^* \leq \d$  and $\Rr(\m^*,\K,\r,\p) = \Rr(\d,\K,\r,\p)$. Then $\S$ is uniquely identifiable from its canonical projections onto all possible tensorized observation patterns of size $\m \geq \m^* + \p$.}
\end{myCorollary}

\new{The key assumption made in Corollary \ref{cor:suffuos} is that $\Rr(\m^*,\K,\r,\p) = \Rr(\d,\K,\r,\p)$. Characterizing the set of values for which this condition holds in all generality appears to be a difficult problem (see, e.g., \cite{carlini2012subspace}). However, using existing results \cite{hartshorne1982droites,derksen,carlini2012subspace} that characterize exact values of $\Rr(\d,\K,\r,\p)$ we can establish the following special cases:}
\begin{myProposition}\label{prop:hilbert}
\new{
Let $\m_0$ be the smallest integer such that {$\binom{\m_0+\p-1}{\p}> \K\binom{\r+\p-1}{\p}$} and set $\m^* = \max\{\m_0,2\r\}$. Then $\Rr(\m^*,\K,\r,\p) = \K\binom{\r+\p-1}{\p} = \Rr(\d,\K,\r,\p)$ in the following cases:
\begin{itemize}
  \item[(a)] $\p=2$, for any $\K$, for any $\r$ (\ie any generic UoS under a quadratic tensorization)
  \item[(b)] $\p\geq 3$, for any $\K$, and $\r=1$ or $2$ (\ie any generic union of 1-D subspaces, or any generic union of 2-D subspaces, under any higher order tensorization)
  \item[(c)] $\p\geq 3$, $\K \leq \p$, for any $\r$ (\ie any generic UoS consisting of at most $\p$, $\r$-dimensional subspaces under a $\p^{\text{th}}$-order tensorization)
\end{itemize}
}
\end{myProposition}
\new{Case (a) is due to \cite[Theorem 3.2]{carlini2012subspace}, case (b) is due to \cite{hartshorne1982droites}, and case (c) is due to \cite[Corollary 4.8]{derksen} (see also \cite[Corollary 2.16]{ma2008estimation}).}

The quantity $\m_0$ defined in Proposition \ref{prop:hilbert} is $O(\K^{1/\p} \r)$, hence so is $\m^* = \max\{2\r,\m_0\}$. Therefore, Proposition \ref{prop:hilbert} combined with Corollary \ref{cor:suffuos} shows that a sampling rate of $\m = O(\K^{1/\p} \r + \p)$ is sufficient for unique identifiability of the tensorized subspace arising from a generic union of $\K$ subspaces of dimension $\r$ (under one of the assumptions (a)-(c) in Corollary \ref{cor:suffuos}). When $\m^* \geq 2\r$, the sampling rate identified in Corollary \ref{cor:suffuos} is only slightly more than the minimal sampling rate $\m_0$ given in \eqref{eq:hypnc} conjectured to be necessary and sufficient for unique identifiability in \cite{ongie}. Specifically, in this case the sampling rate in Corollary \ref{cor:suffuos} is $\m \geq \m_0 + \p$. By the discussion following Corollary \ref{cor:basicnecc_uos}, this rate is also necessary provided $\m_0 > \p$ and provided the ambient dimension $\d$ is sufficiently large. Hence, in these cases, there is only a gap of up to $\p$ observations per column between our necessary and sufficient conditions (\ie $\m \geq \m_0$ versus $\m \geq \m_0 + \p$). 

In general, we conjecture the value of $\m^*$ as defined in Proposition \ref{prop:hilbert} is always sufficient to ensure $\Rr(\m^*,\K,\r,\p) = \Rr(\d,\K,\r,\p)$ for higher order tensorizations $\p \geq 3$. While proving this may be difficult, this condition can also be checked numerically by sampling sufficiently many points from a randomly generated UoS with the specified parameters and computing the rank of tensorized matrix. However, we reiterate that empirical evidence leads us to believe the necessary sampling rate for UoS identified in Corollary \ref{cor:basicnecc_uos} is also sufficient, which generally is less than the rate given in Corollary \ref{cor:suffuos}.

\subsection{Implications for LADMC}
\subsubsection{Sample complexity}\label{sec:sampcomplexity}
The above results are stated in terms of the unique identification of the tensorized subspace $\S$ from canonical projections. However, unique identification of the tensorized subspace also implies unique completion of the tensorized matrix $\XTp$ provided the observation pattern matrix $\O$ contains sufficiently many duplicates of each of the $\binom{\d}{\m}$ possible observation patterns so that all the canonical projections of the tensorized subspace can be determined. 

For example, suppose each of the $\N$ columns in $\O$ is drawn randomly from the $\n = \binom{\d}{\m}$ possible observation patterns of size $\m$ in $\d$ coordinates. Then, by a variant of the coupon collector's problem, with high probability $\O$ will contain $\Rr$ copies of each observation pattern provided the number of columns $\N = \n \log \n + (\Rr-1)\n \log\log \n + O(\n)$, which reduces to $\N = O(\Rr \d^\m \log \d)$. If every subset of $\Rr$ columns of $\X$ is in general position, it will be possible to determine the canonical projections of $\S$ from the columns of $\XTp$. Once $\S$ is recovered, then it is possible to uniquely complete the matrix by projecting each incomplete column of $\XTp$ onto $\S$, and performing the ``de-tensorization'' step of LADMC (Step 4 of Algorithm \ref{ladmcAlg}).

While the above argument establishes a sufficient number of columns to uniquely complete a LAD matrix with high probability, we believe this is a vast overestimate of how many columns are truly necessary and sufficient for successful completion with LADMC. For example, a naive extension of the results in \cite{LRMCpimentel} would lead one to believe that $\N \geq (\Rr+1)(\Dd-\Rr)$ columns are necessary and sufficient for unique recovery of $\S$, which is far less than the estimate given in the previous paragraph. 
However, the tensorization process violates many of the genericity assumptions in \cite{LRMCpimentel}, which prevents the direct extension of these results to the present setting. 
Nevertheless, empirically we observe that LADMC often successfully recovers synthetic variety data with the necessary minimal number observations per column (selected uniformly at random) provided there are $\N = O(\Rr(\Dd-\Rr))$ columns, and we conjecture this is the true necessary and sufficient orderwise number of columns needed for recovery with LADMC (see Figure \ref{fig:mvsp} and Section \ref{sec:Exp} for more discussion on this point).

\subsubsection{Tightness of bounds}
In the special case of a union of $\K$ subspaces of dimension $\r$, Corollary \ref{cor:suffuos} shows that $\m = O(\K^{1/\p}\r+\p)$ observations per data column \edit{is information-theoretically sufficient for unique identification of the tensorized subspace, and hence unique completion of the original matrix (given sufficiently many data columns, and under some restrictions on $\p, \K$ and $\r$).}
In contrast, the information-theoretic requirements for subspace clustering with missing data (SCMD), which is mathematically equivalent to matrix completion under a union of subspaces (UoS) model, to succeed is $\m = \r+1$ observations per data column \cite{infoTheoretic}. 
If $\p = O(1)$, \ie the tensor order is fixed and not allowed to scale with the number of subspaces, this shows that the \edit{information-theoretic} necessary sample complexity of LADMC is order-wise suboptimal by a factor of $\K^{1/\p}$. However, if  the tensor order $\p$  scales with the number of subspaces $\K$ as $\p = O(\log \K)$ then we have $\m = O(\r + \log \K)$, which is nearly orderwise optimal. Nonetheless, even with fixed and low tensor orders (e.g., $\p = 2,3$), empirically we find that LADMC performs equally well or better than most state-of-the-art SCMD methods on UoS data (see Figure \ref{fig:phase}).

\section{Experiments}
\label{sec:Exp}

In the following experiments we demonstrate the performance of the proposed LADMC algorithm (Algorithm \ref{ladmcAlg}) on real and synthetic data having low algebraic dimension, and empirically verify the information-theoretic sampling requirements for LADMC for unions of subspaces data  given in Section \ref{sec:theory}.

\subsection{Implementation details: LADMC, iLADMC, and VMC}\label{sec:implementation} In our implementation of LADMC (Algorithm \ref{ladmcAlg}) we use iterative singular value hard thresholding (ISVHT) algorithm \cite{iht} to perform LRMC in the tensorized domain. The ``de-tensorization'' step of Algorithm \ref{ladmcAlg} is performed by a rank-1 truncated SVD of each reshaped tensorized column, where the sign ambiguity is resolved by matching with the sign of the observed entry having maximum absolute value in each column.

We also test an iterative version of LADMC (iLADMC), where we perform a small number of ISVHT iterations the tensorized domain, map back to the original domain by the columnwise rank-1 SVD de-tensorization step, fill in the observed entries of the matrix, and repeat these steps until convergence. In the experiments below we ran $30$ iterations of ISVHT for iLADMC. Periodically performing the de-tensorization step amounts to a projection onto the space of matrices in the tensorized space with the necessary tensorized structure -- \ie each column is a vector of the form $\bx^\oell$. While we have no theory to show an iterative approach should outperform LADMC, empirically we find that iLADMC converges much faster than LADMC (in terms of the number of ISVHT steps, which is the main computational bottleneck) and succeeds in completing matrices at lower sampling rates than plain LADMC.

In an earlier work \cite{ongie} we introduced an algorithm called variety-based matrix completion (VMC) designed to achieve the same goal as LADMC and iLAMDC. In particular, VMC attempts to minimizes the non-convex Schatten-$\q$ quasi-norm ($0 < \q < 1$) of the tensorized matrix $\X^\oell$ using an iterative reweighted least squares approach \cite{fazel}. The VMC algorithm is most similar to iLADMC, since it also enforces the tensorized structure at each iteration. 

\edit{Each of these approaches has different trade-offs with respect to sample complexity, memory complexity, and computational complexity. In Table \ref{tab:complexity} we compare these complexities for completing data belonging to a union of subspaces. Note that LADMC, iLADMC, and VMC have similar sample complexities, while LADMC/iLADMC have lower memory and computational complexity when the dimension of the data points raised to the tensor order $d^p$ is less than the number of data columns $N$.}

\begin{table}[]
    \centering
    \edit{
    \begin{adjustbox}{width=\columnwidth}
    \begin{tabular}{c|c|c|c|c}
          & LRMC & LADMC &  iLADMC &  VMC  \\
          \hline
          Minimum necessary sampling rate (observations/column)  & $O(Kr)$  & $O(K^{1/p}r)$ & $\leq O(K^{1/p}r)$ & $\leq O(K^{1/p}r)$\\
          \hline
         Memory requirements (bytes) & $O(dN)$ & $O(d^p N)$ & $O(d^pN)$ & $O(N^2)$\\
         \hline
              Computational complexity (flops/iteration)  & $O(dNKr)$  & $O(d^pNKr^p)$ & $O(d^pNKr^p)$ & $O(N^2Kr^p)$
    \end{tabular}
    \end{adjustbox}
    }
    \caption{\edit{Sample, memory, and computational complexity of algorithms for completing a $d\times N$ matrix whose columns belong to a union of $K$ $r$-dimensional subspaces. We compare rates for low-rank matrix completion (LRMC) via the iterative singular value hard thresholding algorithm \cite{iht}, low-algebraic dimension matrix completion (LADMC) via Algorithm \ref{ladmcAlg} and the iterative variant (iLADMC) using a $p$th order tensorization, and the kernel-based variety matrix completion (VMC) algorithm of \cite{ongie}.}}
    \label{tab:complexity}
\end{table}

\subsection{Sample complexity of union of subspaces data}
\begin{figure*}[ht!]
\centering
\includegraphics[width=\textwidth]{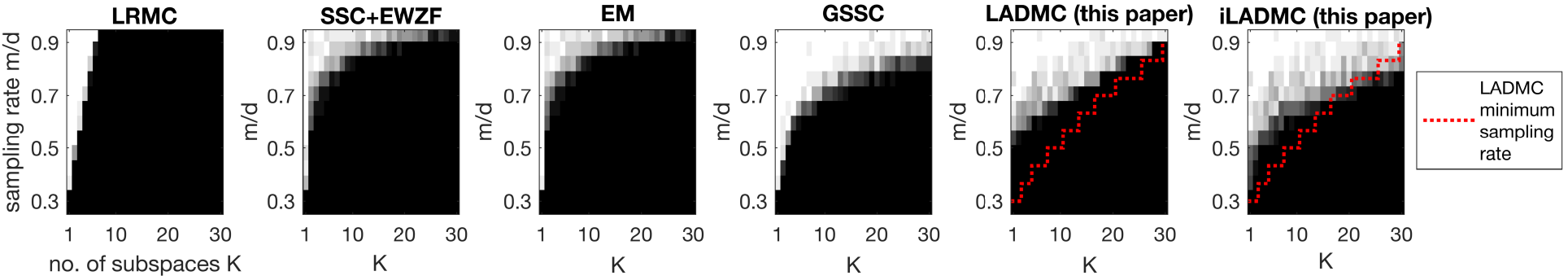}
\caption{\small Phase transitions for matrix completion of synthetic union of subspaces data. We simulated data belonging to $\K$ subspaces and sampled each column of the data matrix at a rate $\m/\d$, and perform matrix completion using LRMC, state-of-the-art subspace clustering based algorithms (SSC+EWZF, GSSC, EM), and the proposed LADMC and iLADMC algorithms with quadratic tensorizations. Grayscale values indicate the fraction of random trials where the matrix were successfully recovered; white is 100\% success and black is 100\% failure. The red dashed line indicates the minimal information-theoretic sampling rate $\m/\d = O(\sqrt{\K})$ needed for LRMC to succeed in the tensorized domain as specified by Corollary \ref{cor:basicnecc_uos}.}
\label{fig:phase}
\end{figure*}

Figure \ref{fig:phase} shows the performance of the LADMC and iLADMC algorithms against competing methods for the recovery of synthetic union of subspaces data with missing entries. We generated $\d\times \N$ data matrices whose columns belong to a union of $\K$ subspaces each of dimension $\r$, and sampled $\m$ entries in each column, selected uniformly at random. We used the settings $\d=15$, $\N=50\K$, $\r=2$, for varying measurements $\m$ and number of subspaces $\K$, and measured the fraction of successful completions over $25$ random trials for each pair $(\m,\K)$. We judged the matrix to be successfully completed if the normalized root mean square error $\|\mathbf{X}-\mathbf{ X_0}\|_F/\|\mathbf{X_0}\|_F$ was less than $10^{-4}$, where $\mathbf{X}$ is the recovered matrix and $\mathbf{X_0}$ is the ground truth matrix and $\|\cdot\|_F$ denotes the Frobenius norm. Here we compared against low-rank matrix completion (LRMC) via iterative singular value hard thresholding (ISVHT) \cite{iht} in the original matrix domain, and three methods based on subspace clustering: sparse subspace clustering (SSC) with entry-wise zero fill \cite{ewzf} followed by LRMC on each identified cluster (SSC+EWZF), the expectation-maximization (EM) algorithm proposed in \cite{ssp14}, and the group-sparse subspace clustering algorithm \cite{GSSC} followed by LRMC on each cluster (GSSC). The subspace clustering algorithms were passed the exact rank and number of subspaces. The EM and GSSC algorithms were initialized with the subspace clustering obtained by SSC-EWZF. Any remaining free parameters in these algorithms were set via cross-validation. For LADMC and iLADMC we used a quadratic tensorization ($p=2$) and LRMC steps for these algorithms were performed via ISVHT with the rank threshold parameter set to the exact rank of the tensorized matrix.

\begin{figure}
    \begin{minipage}{0.45\textwidth}\centering\includegraphics[height=4.7cm]{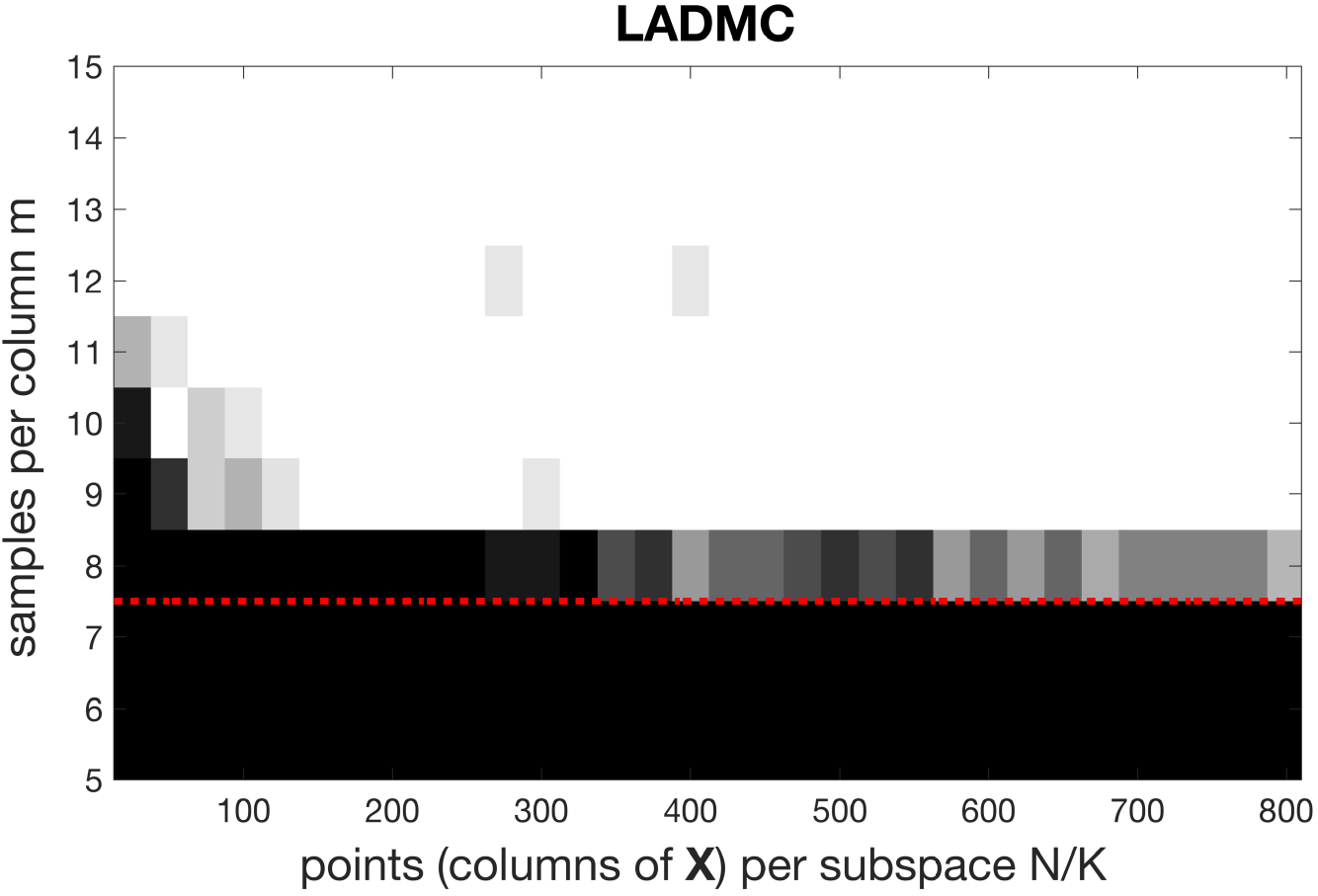}\\[0.5em]
    \quad\includegraphics[height=0.5cm]{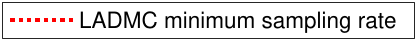}
\caption{\small Effect of number of data columns per subspace $\N/\K$ on probability of exact LADMC recovery of synthetic UoS data ($\K=10$ subspaces of dimension $\r = 2$ in $\d=15$ dimensional space). As the number of columns per subspace increases the probability of exact recovery is approaching 1 for $\m \geq 8$, the necessary minimum number of samples per column identified by Corollary \ref{cor:basicnecc_uos}.}
\label{fig:mvsp}
    \end{minipage}
    \hfill
    \begin{minipage}{0.50\textwidth}\centering\includegraphics[height=4.7cm]{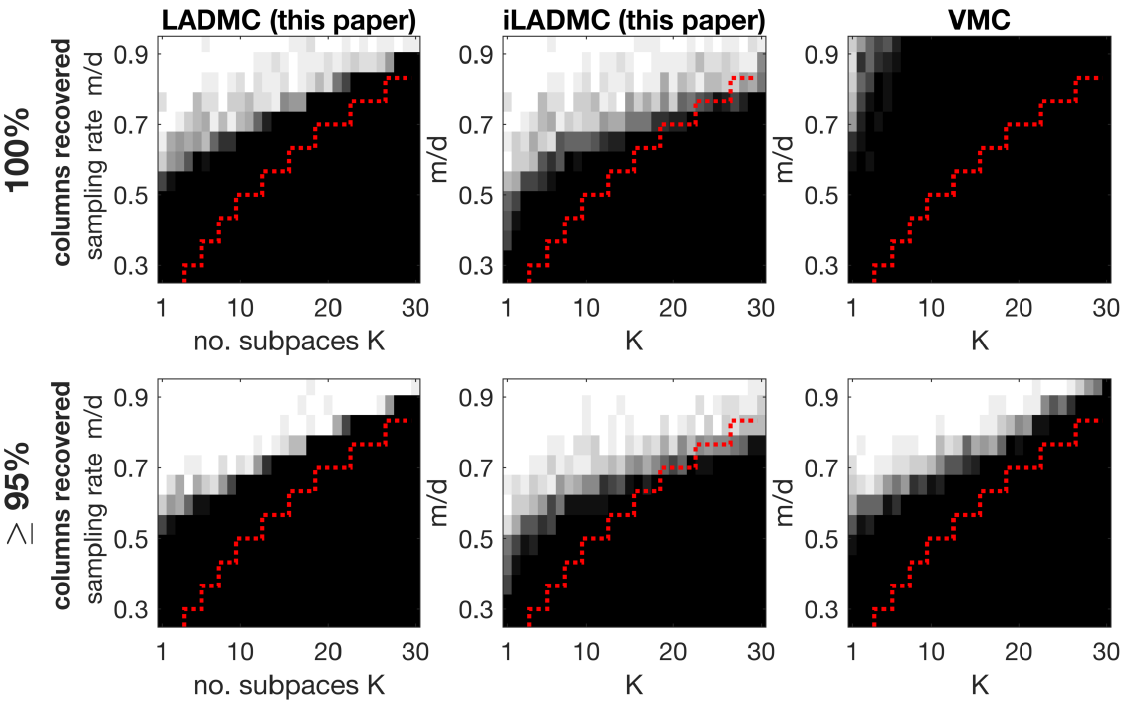}\\[0.5em]
    \quad\includegraphics[height=0.5cm]{Figures/legend_only_hor.pdf}
\caption{\small Relative performance of LADMC and iLADMC compared with VMC \cite{ongie} for recovery of synthetic UoS data. LADMC and iLADMC succeed with high probability at recovering all the columns where VMC often fails (top row). The algorithms perform similarly when comparing the probability of recovering at least $95\%$ columns (bottom row).}
\label{fig:vmc_compare}
\vspace{2em}
    \end{minipage}
\end{figure}

We find that LADMC is able to successfully complete the data when the number of measurements per column in the tensorized domain exceeds the information-theoretic bounds established in Corollary \ref{cor:basicnecc_uos}, as indicated by the red dashed line in Figure \ref{fig:phase}. This is a substantial extension over standard LRMC: for these settings, LADMC is able to complete data matrices drawn from up to $\K=30$ subspaces, whereas LRMC is limited to data drawn from less than $\K=7$ subspaces. However, for LADMC there is a small gap between the information-theoretic bound and the true phase transition, which is most apparent where the number of subspaces and sampling rate is low (lower-left of the plot), but this gap closes as the number of subspaces and sampling rate increases (upper-right of the plot). We hypothesize this is due to insufficiently many data columns (see Figure \ref{fig:mvsp} and the discussion below). This gap is less pronounced for iLADMC, and in fact, in the upper-right of the plot iLADMC shows recovery below the LADMC information-theoretic bound. We conjecture this is because iLADMC is enforcing extra nonlinear constraints that are not accounted for in our theory, which may reduce the sample complexity relative to non-iterative LADMC, both in terms of necessary number of data columns and the number of samples per column. We also observe that the performance of LADMC and iLADMC is competitive with the best performing subspace clustering-based algorithm, which in this case is GSSC. 

In Figure \ref{fig:mvsp} we investigate the effect of the number of data columns per subspace in the overall recovery performance of LADMC for synthetic UoS data. Here we use the same settings as in the previous experiment, but fix the number of subspaces to be $\K=10$ and vary the number of columns per subspace, $\N/\K$, and the number of random measurements per column, $\m$. In this case, the tensorized subspace has rank $\Rr = 30$ and the necessary minimum number of observations per column according to Corollary \ref{cor:basicnecc_uos} is $\m = 8$. Observe that as the number of columns per subspace increases, the probability of exact recovery is approaching one for $\m \geq 8$, as predicted by Corollary \ref{cor:basicnecc_uos}. The minimum number columns per subspace needed for exact recovery we conjecture to be $\N/\K =O(\Rr(\Dd-\Rr)/\K)$ (see Section \ref{sec:sampcomplexity}). Assuming the constant in the order-wise expression to be one, we have $\N/\K \approx 270$. Note that we do see exact recovery at $\m = 9$ samples per column when $\N/\K = 270$ and partial success at $\m = 8$ with two- or three-fold more columns, as predicted.

\subsection{Comparison with VMC}

In Figure \ref{fig:vmc_compare} we compare the relative performance of LADMC and iLADMC with VMC for the same synthetic unions of subspaces data as in Figure \ref{fig:phase}. One drawback of VMC observed in \cite{ongie} is that it often failed to complete a small proportion of the data columns correctly, even at high sampling rates on synthetic data. Consistent with the results in \cite{ongie}, we find that VMC and LADMC/iLADMC perform similarly when comparing probability of recovering at least $95\%$ columns. However, LADMC and iLADMC both recover $100\%$ of the data columns correctly above the minimum sampling rate, whereas VMC mostly fails under this more strict recovery criterion. This shows that LADMC/iLADMC could have some empirical benefits over VMC if high accuracy solutions are desired.

\subsection{Higher order tensorizations}\label{sec:hot}
\begin{figure}[ht!]
\centering
\includegraphics[width=0.5\columnwidth]{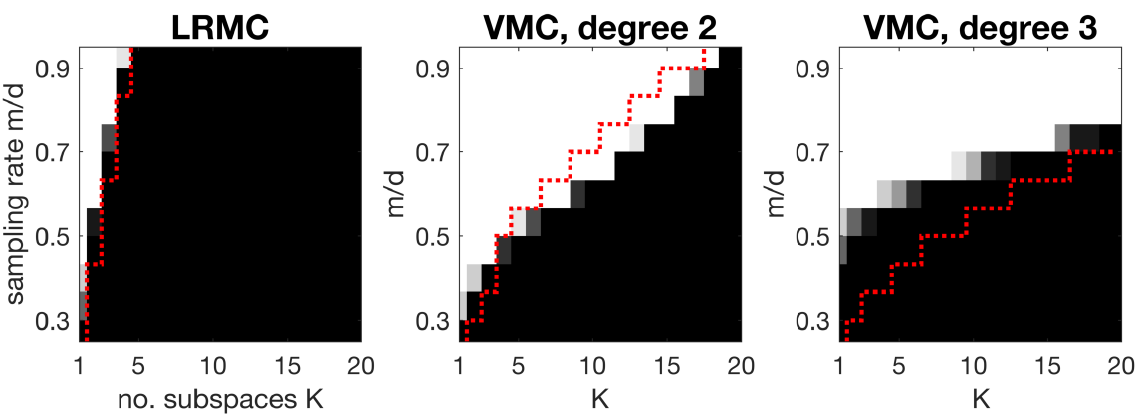}\\
\quad\includegraphics[width=0.35\columnwidth]{Figures/legend_only_hor.pdf}
\caption{\small Phase transitions for matrix completion for unions of subspaces using no tensorization (LRMC), 2nd order tensorization (VMC, degree 2), and 3rd order tensorizaiton (VMC, degree 3). The phase transition follows closely the LADMC minimum sampling rate established in Corollary 1, which is $\m = O(\K^{1/\p})$ where $\K$ is the number of subspaces and $\p$ is the tensor order. Here the ambient dimension is $\d = 15$ and the subspace dimension is $\r = 3$. (Figure adapted from \cite{ongie}).}
\label{fig:higherorder}
\end{figure}

In Figure \ref{fig:higherorder} we experimentally verify the predicted minimal sampling rate for UoS data with higher order tensorizations specified in Corollary \ref{cor:basicnecc_uos}. In this work we do not pursue higher order $\p\geq 3$ LADMC with Algorithm \ref{ladmcAlg}, due to poor scalability with respect to the ambient dimension $\d$ and a lack of an efficient implementation of the de-tensorization step, which prohibited us from investigating the phase transition behavior of LADMC over a reasonable range of the number of subspace $\K$. Instead, we verify our predictions using VMC algorithm \cite{ongie}, for which the sufficient conditions of Corollary \ref{cor:suffuos} also hold (although the necessary conditions of Corollary \ref{cor:basicnecc_uos} may not hold). We find that the phase transition recovery follows the dependence $\m = O(\K^{1/\p})$ for tensor orders $p=2,3$ as predicted by Corollaries \ref{cor:basicnecc_uos} and \ref{cor:suffuos}.

\subsection{Experiments on real data}
\begin{figure}
\centering
\begin{minipage}{0.45\columnwidth}
\begin{tabular}{ccccc}
\textsf{Truth} & \textsf{Samples} & \textsf{LRMC} & \textsf{LADMC} & 
\end{tabular}
\centering
\includegraphics[width=\textwidth]{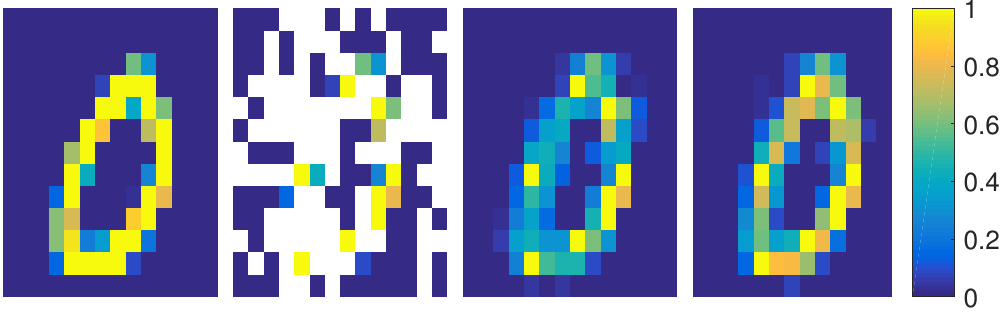}\\[0.2em]
\includegraphics[width=\textwidth]{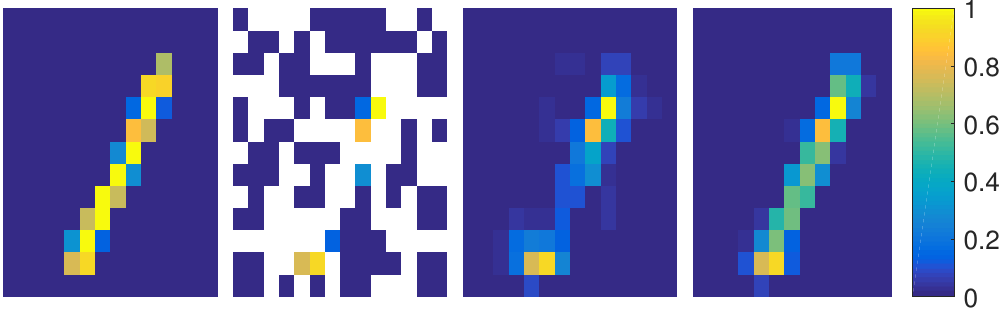}
\end{minipage}
~
\begin{minipage}{0.45\columnwidth}
\begin{tabular}{ccccc}
\textsf{Truth} & \textsf{Samples} & \textsf{LRMC} & \textsf{LADMC} & 
\end{tabular}
\centering
\includegraphics[width=\textwidth]{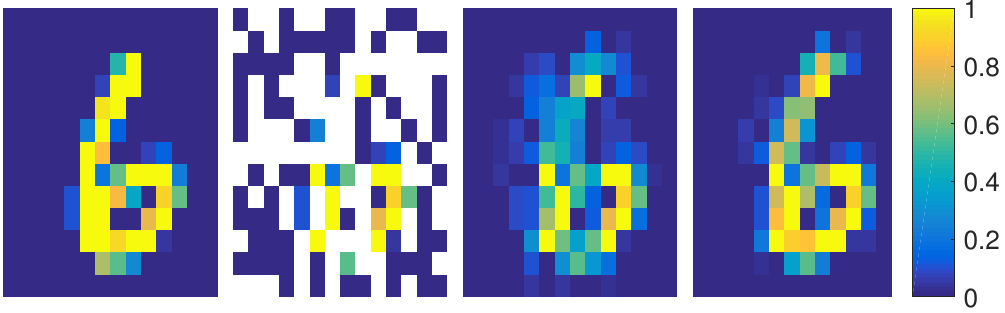}\\[0.2em]
\includegraphics[width=\textwidth]{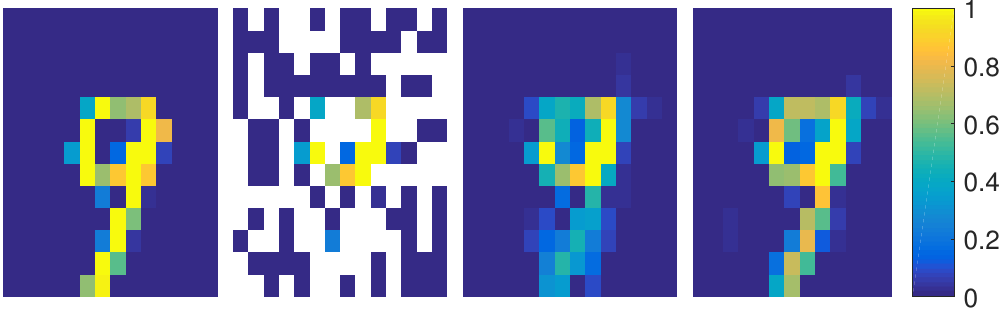}
\end{minipage}
\caption{\small Representative examples of matrix completion on MNIST dataset. Here we randomly remove $50\%$ of the pixels in each MNIST image and attempt to jointly recover the missing pixels of all images by low-rank matrix completion (LRMC) and low algebraic dimension matrix completion (LADMC) using a quadratic tensorization $(\p = 2)$.}
\label{fig:mnist}
\end{figure}

Here we illustrate the performance of LADMC and iLADMC on three real world datasets\footnote{Available online: Oil Flow \url{http://inverseprobability.com/3PhaseData.html}, Jester-1  \url{http://goldberg.berkeley.edu/jester-data/}, MNIST  \url{http://yann.lecun.com/exdb/mnist/}. For computational reasons, we reduced the size of the MNIST dataset by selecting a random subset of 20,000 images and downsampling each image by a factor of two in both dimensions.}: the Oil Flow dataset introduced in \cite{oilflow}, the Jester-1
recommender systems dataset  \cite{jester}, 
and the MNIST digit recognition dataset introduced in \cite{lecun1998gradient}.
We chose these datasets to demonstrate the feasibility of LADMC on a variety of data sources, and because they had sufficiently small row dimension for LADMC/iLADMC to be computationally practical. For the Oil Flow and MNIST datasets we simulate missing data by randomly subsampling each data column uniformly at random, using a 50\%-25\%-25\% training-validation-test split of the data. For the Jester-1 dataset we used 18 randomly selected ratings of each user for training, 9 randomly selected ratings for validation and the remainder for testing. As baselines we compare with filling the missing entries with the mean of the observed entries within each column (Mean-fill), and with LRMC via nuclear norm minimization \cite{recht2010guaranteed}, which outperformed LRMC via singular value iterative hard thresholding \cite{iht} on these datasets. For the LRMC routine within LADMC we set the rank cutoff $\Rr$ to the value that gave the smallest completion error on the validation set, and use the same rank cutoff $\Rr$ for iLADMC. For all methods we report the root mean square error (RMSE) of the completion on the test set. We find that LADMC/iLADMC gives significantly lower RMSE on the Oil Flow and MNIST datasets relative to the baselines; iLADMC gives lower RMSE than LADMC on the Oil Flow dataset, but performs similarly to LADMC on the others. Figure \ref{fig:mnist} illustrates the improvement of LADMC over LRMC on a selection of examples from the MNIST dataset. We see less differences between LRMC and LADMC/iLADMC on the Jester-1 dataset, where LADMC/iLADMC give nearly the same RMSE as LRMC. Because of lower sampling rate for the Jester-1 dataset, the rank cutoff $\Rr$ in LADMC was kept small to avoid overfitting, and we suspect in this case LADMC is fitting a linear subspace to the data, which would explain the similar performance to LRMC.

\begin{table}
\centering
\begin{adjustbox}{width=0.7\columnwidth}
\begin{tabular}{c|c|c|c|c|c|c}
& & & \multicolumn{4}{c}{ Completion RMSE on Test Set} \\
Dataset & Size & Samples & Mean-fill&  LRMC & LADMC & iLADMC \\ \hline
Oil flow & 12$\times$1000 & 50\% & 0.237 & 0.164 &  0.155 & \bf 0.127\\ \hline
Jester-1 & 100$\times$24983 & 18\% & 4.722 & \bf 4.381 & 4.420 & 4.394\\ \hline
MNIST & 196$\times$20000 & 50\% & 0.309 & 0.210 & \bf 0.187 & \bf 0.187 \\ \hline
\end{tabular}
\end{adjustbox}
\vspace{0.5em}

\caption{Matrix completion results on real data}
\label{table:realdata}
\end{table}

\section{Conclusion}
The theory and algorithms presented in this paper give new insight into conducting matrix completion when the matrix columns correspond to points on a nonlinear algebraic variety, including union of subspaces as a special case. Unlike most matrix completion methods assuming a union of subspace model, the proposed approach does not necessitate an intermediate subspace clustering step that can be fragile in the presence of missing data. 

The theoretical guarantees in this work focus on unique identifiability of the tensorized subspace from canonical projections -- \ie we assume we observe multiple columns with each possible observation pattern. This assumption is not always met in practice, yet the proposed LADMC algorithm nevertheless performs well empirically. An important avenue for future study are conditions for unique completion of partially sampled data matrices.

In the experimental portion of this work we primarily focused on LADMC with a quadratic tensorization. Yet, we also showed our approach and results generalize to LADMC with higher-order tensorizations. In principle, this extension would facilitate the completion of data belonging to a richer class of varieties and with more missing data. However, the computational complexity of LADMC scales as $O(\d^\p)$, where $\d$ is the ambient (data) dimension and $\p$ is the tensor order, making our approach computational challenging for even modest data dimensions $\d$.

One potential solution is to use a kernelized algorithm, such as VMC \cite{ongie}, that avoids explicitly constructing a large scale tensorized matrix. Unfortunately, kernelized approaches have complexity and storage requirements that scale quadratically with the number of data columns, making such an approach computationally challenging for large datasets.

We are actively investigating memory and computationally efficient algorithms that allow more efficient extensions of the LADMC approach for higher-order tensorizations. Along these lines, recent work investigates efficient online algorithms for a class of nonlinear matrix completion problems that includes the LADMC model \cite{fan2019online}. \edit{An alternative approach would be to incorporate matrix sketching techniques to allow for more efficient computation of the SVD in the tensorized domain. For example, recent work that focuses on  sketches for matrices with Kronecker product structure, such as TensorSketch \cite{pham2013fast} and its modifications \cite{ahle2020oblivious} or the Kronecker Fast Johnson-Lindenstrauss Transform \cite{jin2019faster}, could potentially be adapted to our setting.}

\appendix
\section{Proof of Theorem \ref{thm:suff}}\label{sec:app:thm1}
We prove Theorem \ref{thm:suff} by showing we can construct an observation pattern matrix $\Ups^* \in \{0,1\}^{\Dd\times (\Dd-\Rr)}$ such that the resulting \emph{constraint matrix} $\bA = \bA(\S,\Ups^*) \in \R^{\Dd\times (\Dd-\Rr)}$ satisfies Lemma \ref{lem1}, \ie $\dim \ker \bA^\T = \Rr$.

Note that in the tensor domain we observe projections of the tensorized subspace onto subsets of $\M = \binom{\m+\p-1}{\p}$ coordinates where $\M$ may be larger than $\Rr+1$. However, from any canonical projection of the tensorized subspace onto $\M > \Rr+1$ coordinates we can also recover its canonical projections onto any subset of $\Rr+1$ coordinates of the $\M$ coordinates. That is, if we observe one canonical projection $\S_\ups$ with $|\ups| = \M$ then we also have access to all canonical projections $\S_{\ups'}$ where $\ups'$ is any observation pattern with $\text{supp}(\ups') \subset \text{supp}(\ups)$ and $|\ups'| = \Rr+1$. 

To express this fact more succinctly, we introduce some additional notation. For any observation pattern matrix $\Ups \in \{0,1\}^{\Dd\times\n}$ whose columns all have greater than $\Rr$ nonzeros, let $\widehat{\Ups}$ denote the matrix of observation patterns having exactly $\Rr+1$ non-zeros that can be generated from the observation patterns in $\Ups$. For example, if $\Dd = 4$, $\Rr = 1$, then from the two $3$-dimensional projections indicated in $\Ups$ below we obtain the five $2$-dimensional projections indicated in $\widehat{\Ups}$ below:
\begin{equation*}
\Ups = 
\begin{bmatrix}
1 & 0\\
1 & 1\\
1 & 1\\
0 & 1
\end{bmatrix}
\mapsto \widehat{\Ups} = 
\begin{bmatrix}
1 & 0 & 1 & 0 & 0\\
1 & 1 & 0 & 0 & 1\\
0 & 1 & 1 & 1 & 0\\
0 & 0 & 0 & 1 & 1
\end{bmatrix}.
\end{equation*}

Recall that every coordinate in tensor space is associated with an ordered tuple $(\k_1,...,\k_\p)$ satisfying $1 \leq \k_1 \leq \cdots \leq \k_\p \leq \d$, where each $\k_i$ indicates one of the $\d$ coordinates in the original space. 
We assume that coordinates in tensor space are ordered such that for all $\m =1,...,\d$, the first $\M = \binom{\m+\p-1}{\p}$ coordinates correspond to all tuples $1\leq \k_1 \leq ... \leq \k_d \leq m$.  We call this the \emph{standard ordering.}

We now show that if $\Ups$ consists of all tensorized observation patterns of a certain size then the expanded observation pattern matrix $\widehat{\Ups}$ contains several submatricies having a useful canonical form.

\begin{myLemma}\label{lem:goodsamp}
Fix a tensor order $\p \geq 2$. Suppose the columns of $\Ups$ are given by all $\binom{\d}{\m}$ tensorized observation patterns of size $\m \geq \m^*+\p$ where $\m^*$ is the smallest integer such that $\M^* := \binom{\m^*+\p-1}{\p} > \Rr$, and let $\widehat{\Ups}$ be its expanded observation pattern matrix having exactly $\Rr+1$ ones per column. Then any permutation of the first $\M^*$ rows of $\widehat{\Ups}$ has a submatrix of the form
\begin{align}
\label{goodSamplingEq}
\Ups^\star \ = \ \left[ \begin{array}{c}
 \Scale[1.5]{\bs{1}} \\ \hline
\\
\Scale[1.5]{\I} \\ \\
\end{array}\right]
\begin{matrix}
\hspace{-.7cm} \left. \begin{matrix} \\ \end{matrix} \right\} \Rr \\
\left. \begin{matrix} \\ \\ \\ \end{matrix} \right\} \Dd-\Rr.
\end{matrix}
\end{align}
where ${\bs{1}}$ is the $\Rr \times (\Dd-\Rr)$ matrix of all ones, and $\I$ is the $(\Dd-\Rr)\times(\Dd-\Rr)$ identity.
\end{myLemma}
\begin{proof}

Let $\ups^*_\j$ denote the $\j{\text{th}}$ column of $\Ups^\star$ in \eqref{goodSamplingEq}, whose first $\Rr$ entries are ones and $(\Rr+\j)^{\rm th}$ entry is one while the rest of its entries are zero. We show that $\ups^*_\j$ is  a column $\widehat{\ups}$ in the expanded matrix $\widehat{\Ups}$. Let $(\k_1,...,\k_\p)$ be the ordered tuple corresponding to the $(\Rr+\j)^{\rm th}$ coordinate in the tensor space.
Note that a column $\ups$ in $\Ups$ has nonzero $(\Rr+\j)^{\rm th}$ entry if and only if the corresponding observation pattern $\o$ is nonzero in entries $\k_1,\k_2,...,\k_\p$.
Let $\o$ be any column of $\O$ such that all entries indexed by $\{1,...,\m^*,\k_1,...,\k_\p\}$ are equal to one. By construction $\ups = \o^\oell$ has ones in its first $\M^*$ entries and must also have a one at the $(\Rr+\j){\text{th}}$ entry. This shows that $\ups^*_\j$ is a column of the expanded matrix $\widehat{\Ups}$, and thus of any permutation of the first $\M^*$ rows of $\widehat{\Ups}$. Since this is true for every $\j=1,\dots, \Dd-\Rr$, we know that $\Ups$ will produce a matrix $\widehat{\Ups}$ containing $\Ups^\star$ as in \eqref{goodSamplingEq} (and likewise for any permutation of the first $\M^*$ rows of $\widehat{\Ups}$).
\end{proof}

Now we are ready to give the proof of Theorem \ref{thm:suff}.

\begin{proof}[Proof of Theorem \ref{thm:suff}]
First we will permute the tensorized coordinate system into a convenient arrangement. Assume there exists an tensorized observation pattern $\ups = \o^\oell$ such that $\dim \S_\ups = \Rr$. Define $\m^* = |\o|$. Without loss of generality, we may permute coordinates in the original domain such that the first $\m^*$ entries of $\o$ are ones. Under the standard ordering of tensor coordinates, this means the first $\M^* = \binom{\m^*+\p-1}{\p}$ entries of $\ups = \o^\oell$ are ones. Since $\dim \S_\ups = \Rr$, there exists an observation pattern $\ups'$ with $\supp(\ups')\subset \supp(\ups)$ having exactly $\Rr$ ones such that $\dim \S_{\ups'} = \Rr$. We may permute the first $\M^*$ coordinates in tensor space so that $\ups'$ has all its ones in the first $\Rr$ coordinates. Thus, the restriction of $\S$ to the first $\Rr$ coordinates is $\Rr$-dimensional (\ie $\dim \S_{\ups'} = \Rr$).

Now suppose we observe canonical projections of $\S$ onto all tensorized observation patterns of size $\m \geq \m^* + \p$, which we collect into a matrix $\Ups \in \{0,1\}^{\Dd\times \binom{\d}{\m}}$.  Then by Lemma \ref{lem:goodsamp} there exists a submatrix $\Ups^* \in \{0,1\}^{\Dd \times (\Dd-\Rr)}$ of the expanded matrix $\widehat{\Ups}$ having the form \eqref{goodSamplingEq}. Hence, from canonical projections of $\S$ onto observation patterns in $\Ups$ we can derive all canonical projections of $\S$ onto observation patterns in $\Ups^*$. 

For $j=1,\dots,\Dd-\Rr$, let $\ups_j^*$ be the $j$th column of $\Ups^*$. Since $\ups_j^*$ has exactly $\Rr+1$ ones, and the restricted subspace $\S_{\ups_j^*}$ is at most $\Rr$ dimensional, the orthogonal complement of the restricted subspace $\S_{\ups_j^*}$ is positive dimensional, and so there exists at least one non-zero constraint vector $\a_j \in (\S_{\ups_j^*})^\perp$. Let $\a_j^* \in \R^d$ be the vector whose restriction to $\ups_j$ is equal to $\a_j$ and zero in its other entries. Then consider the constraint matrix $\bA^* = [\ba_1^*,...,\ba_{\Dd-\Rr}^*]$, which has the same dimensions as $\Ups^*$ and is such that an entry of $\bA^*$ is nonzero only if the corresponding entry of $\Ups^*$ is nonzero. In particular, this means that
\begin{equation}
  \bA^* = 
  \begin{bmatrix}
  \bA_0^* \\ \bA_1^*
  \end{bmatrix} \in \R^{\Dd\times (\Dd-\Rr)}
\end{equation}
where $\bA_1^* \in \R^{(\Dd-\Rr) \times (\Dd-\Rr)}$ is a diagonal matrix. To finish the proof it suffices to show the diagonal entries of $\bA_1^*$ are all nonzero, since this would imply $\rnk(\bA^*) = \Dd-\Rr$, and hence $\dim \ker [(\bA^*)^\T] = \Rr$, which by Lemma \ref{lem1} implies the subspace $\S$ is uniquely identifiable.

Showing the diagonal entries of $\bA_1^*$ are all nonzero is equivalent to showing the constraint vector $\ba_j^*$ is non-zero at entry $(\Rr+\j)$ for all $j=1,...,\Dd-\Rr$. Suppose, by way of contradiction, that $\ba_j^*$ were zero at entry $(\Rr+\j)$. This means that the nonzero support of $\ba_j^*$ is contained in the first $\Rr$ coordinates.
Let $\bB \in \R^{\Dd\times \Rr}$ be a basis matrix for the tensorized subspace $\S$, $\ups'$ be the $\Dd\times 1$ vector with first $\Rr$ rows equal to $1$ and the remainder equal to $0$, and $\bB_{\ups'} \in \R^{\Rr\times \Rr}$ be the matrix composed of the first $\Rr$ rows of $\bB$. By definition $\ba_j^* \in \ker \bB^\T$, and so $\bB^\T\ba^*_j = (\bB_{\ups'})^\T(\ba^*_j)_{\ups'} = \boldsymbol{0}$. Since $\ba^*_j\neq 0$ by definition and because the non-zero support of $\ba^*_j$ is the same as the non-zero support of $(\ba^*_j)_{\ups'}$ by hypothesis, we have $(\ba^*_j)_{\ups'} \neq \boldsymbol{0}$. This implies $(\bB_{\ups'})^\T \in \R^{\Rr\times \Rr}$ is rank deficient, hence so is $\bB_{\ups'}$, or equivalently, $\dim \S_{\ups'} < \Rr$, which is a contradiction. Hence $\ba_j^*$ is non-zero at entry $(\Rr+\j)$ for all $j=1,...,\Dd-\Rr$, and so $\bA_1^*$ is nonzero at every entry along its diagonal, which completes the proof.
\end{proof}

\section{Unique identification of the tensorized subspace in Example \ref{ex:counter2}}\label{sec:supp}
Here we prove that in the special case of a generic union of two 2-D subspaces in $\R^4$ (\ie $\K =2, \r = 2,\d=4$), the corresponding quadratic tensorized subspace is uniquely identifiable from its canonical projections onto all tensorized observation patterns of size $\m = 3$.

Let $\sU, \sV \subset\R^4$ both be two-dimensional subspaces. Consider the quadratic tensorized subspace $\S = \spn\{\x^\otwo : \x \in \sU \cup \sV\} \subset \R^{10}$. Provided $\sU\cap\sV = \{\boldsymbol{0}\}$, one can show $\Rr := \dim \S = 6$ (see  \cite[Example C.32]{gpcabook}).
Suppose we observe canonical projections of $\S$ under all tensorized observation patterns $\Ups = \O^{\otwo} \in \{0,1\}^{10\times 4}$ where
$\O$ is the matrix of all possible observation patterns of size $\m = 3$ in the original domain:
\begin{equation*}
    \O = 
    \begin{bmatrix}
    1 & 1 & 1 & 0\\
    1 & 1 & 0 & 1\\
    1 & 0 & 1 & 1\\
    0 & 1 & 1 & 1\\
    \end{bmatrix},
\end{equation*}
and let $\bA = \bA(\S,\Ups)$ be the corresponding constraint matrix (see Definition \ref{def:cmatrix}).

Recall that the condition $\ker \bA^\T = \Rr = 6$ is necessary and sufficient for unique identification of the tensorized subspace (see Lemma \ref{lem1}). We show this condition is equivalent to the non-vanishing of a polynomial defined in terms of the entries of basis matrices $\bU \in \R^{4\times 2}$ and $\bV \in \R^{4\times 2}$ whose columns span $\sU$ and $\sV$, respectively. This will show that the condition holds generically, \ie for almost all choices of basis matrices $\bU$ and $\bV$.

First, we show that a constraint matrix $\bA$ can be constructed so that its entries are polynomials in entries of $\bU$ and $\bV$. Let $\o$ be any column of $\O$ above, and let $\ups = \o^\otwo$ denote the corresponding column in $\Ups$. Recall that the columns are $\bA$ are built from vectors in the orthogonal complement of the canonical projection $\S_\ups$. In this case $\S_\ups$ is equal to the tensorized subspace arising from from restriction of the UoS to coordinates in $\o$, \ie $\S_{\ups} = \spn \{\y^\otwo : \y\in \sU_\o \cup \sV_\o \}$, where subspaces $\sU_\o\subset \R^3$ and $\sV_\o\subset \R^3$ denote the restrictions of $\sV$ and $\sU$ to coordinates in $\o$, respectively. If we write $\bU = [\bu_1~\bu_2]$ and $\bV = [\bv_1~\bv_2]$, then a basis for $\sU_\o$ and $\sV_\o$ is given by $\U_{\o} = [(\bu_1)_\o~(\bu_2)_\o] \in \R^{3\times 2}$ and $\bV_\o = [(\bv_1)_\o~(\bv_2)_\o] \in \R^{3\times 2}$, respectively. Define vectors $\bu_\o^\perp = (\bu_1)_\o \times (\bu_2)_\o$ and $\bv_\o^\perp = (\bv_1)_\o \times (\bv_2)_\o$ where $\times$ denotes the cross product. In particular, $\bu_\o^\perp$ and $\bv_\o^\perp$ are in the orthogonal complement of $\sU_\o$ and $\sV_\o$, respectively.  Then using basic properties of Kronecker products one can show the vector $\ba_\ups := \bu_\o^\perp \otimes \bv_\o^\perp + \bv_\o^\perp \otimes \bu_\o^\perp$ is in the orthogonal complement of $\S_\ups$. Finally, note that the entries of the vector $\ba_\ups$ correspond to the non-zero entries of the column of $\bA$ associated with the observation pattern $\ups$. Since entries of $\ba_\ups$ are polynomials in the entries of the basis matrices $\bU$ and $\bV$, this shows that all non-zero entries of $\bA$ are polynomials in the entries of the basis matrices $\bU$ and $\bV$.

Now we show that $\dim \ker \bA^\T = \Rr = 6$ holds generically. For this it suffices to show that $\bA \in \R^{10\times 4}$ is generically full rank. To do so, we show $\bA$ has a nonsingular $4\times 4$ submatrix. Recall that the rows of $\bA$ correspond to coordinates in tensor space, \ie all possible pairwise products of coordinates $(x_1,x_2,x_3,x_4) \in \R^4$. Let $\bA'$ be the $4\times 4$ submatrix of $\bA$ obtained by restricting to rows corresponding to tensor coordinates $x_1^2,x_2^2,x_3^2,x_4^2$, and let $P(\bU,\bV)$ denote the determinant of $\bA'$. Note that $P(\bU,\bV)$ is a polynomial defined in terms of the entries of $\bU$ and $\bV$. Hence, to prove the claim it suffices to show $P(\bU,\bV)$ is a polynomial that is not identically zero, since then $P(\bU,\bV) \neq 0$ for almost all choices of basis matrices $\bU,\bV\in\R^{4\times 2}$.  

While it is possible to compute an explicit expression for $P(\bU,\bV)$ this is not necessary since all we need to show is that this polynomial is not identically zero. Instead we show that specific evaluations of this polynomial are nonzero. In particular, define $\bU(a,b)$ and $\bV(c,d)$ by
\begin{equation*}
    \bU(a,b) = 
    \begin{bmatrix}
    1 & 1\\
    a & b\\
    a^2 & b^2\\
    a^3 & b^3
    \end{bmatrix},
    ~~
        \bV(c,d) = 
    \begin{bmatrix}
    1 & 1\\
    c & d\\
    c^2 & d^2\\
    c^3 & d^3
    \end{bmatrix}
\end{equation*}
for any scalars $a,b,c,d \in \R$. Then, plugging these choices into an expression for $P$ obtained via a computer algebra system, we obtain
\begin{equation*}
    P(\bU(a,b),\bV(c,d)) = 16 a^2 b^2 c^2 d^2 (a-b)^4 (c-d)^4 (b c-a d)^2 (a c-b d)^2
\end{equation*}
which is clearly non-zero for appropriate choices of $a,b,c,d$. Hence, $P$ is a non-zero polynomial, and the sufficient condition $\dim \ker \bA^\T = \Rr$ holds generically.

\section*{Acknowledgments}
The authors would like to thank Manolis Tsakiris for his helpful feedback on an early draft of this paper. R.W., R.N. and G.O. were supported in part by AFOSR FA9550‐18‐1‐0166, 
DOE DE‐AC02‐06CH11357,
NSF OAC‐1934637 and NSF
DMS‐1930049. L.B. was supported in part by ARO W911NF1910027, NSF CCF-1845076 and IIS-1838179, and the IAS Charles Simonyi Endowment.

\bibliographystyle{siam}
\bibliography{references}

\end{document}